\documentclass{article}
\usepackage{fullpage}
\usepackage{appendix}
\usepackage[sort, authoryear]{natbib}
\usepackage{fancyhdr}
\usepackage[format=plain,
            labelfont={it,it},
            textfont=it]{caption}
\usepackage[stable, bottom, flushmargin]{footmisc}
\usepackage{graphicx}
\usepackage{xcolor}    
\usepackage{breakcites}

\definecolor{lightblue}{RGB}{0,102,204}
\usepackage[pagebackref, colorlinks=true, linkcolor=lightblue, citecolor=lightblue, urlcolor=lightblue]{hyperref}

\usepackage{pgfplots}
\pgfplotsset{compat = newest}
\usepackage{wrapfig}

\usepackage{amsmath,amsfonts}
\usepackage[all]{xy}
\usepackage{amsthm,stmaryrd}
\usepackage{mathrsfs}

\usepackage{enumerate}
\usepackage{xparse, expl3}
\usepackage{indentfirst}
\usepackage{float}
\usepackage{tikz}
\usepackage{tikzscale}
\usetikzlibrary{arrows.meta,arrows}
\usetikzlibrary{decorations.pathreplacing}
\usetikzlibrary{positioning,chains,fit,shapes,calc,shapes.misc}

\usepackage[capitalize]{cleveref}

\newtheorem{theorem}{Theorem}[section]
\newtheorem{proposition}[theorem]{Proposition}
\newtheorem{lemma}[theorem]{Lemma}
\newtheorem{corollary}[theorem]{Corollary}
\theoremstyle{definition}
\newtheorem{definition}[theorem]{Definition}

\newtheorem{remark}[theorem]{Remark}

\newtheorem{conjecture}[theorem]{Conjecture}

\crefname{theorem}{Theorem}{Theorems}
\crefname{proposition}{Proposition}{Propositions}
\crefname{lemma}{Lemma}{Lemmas}
\crefname{corollary}{Corollary}{Corollaries}
\crefname{definition}{Definition}{Definitions}
\crefname{example}{Example}{Examples}
\crefname{remark}{Remark}{Remarks}
\crefname{algorithm}{Algorithm}{Algorithms}
\crefname{equation}{Equation}{Equations}
\crefname{section}{Section}{Sections}
\crefname{subsection}{Section}{Sections}
\crefname{conjecture}{Conjecture}{Conjectures}

\usepackage{microtype}
\usepackage{hyperref}
\usepackage{graphicx}

\usepackage{bbm}
\usepackage{xfrac}

\makeatletter
\newcommand{\customlabel}[2]{%
   \protected@write \@auxout {}{\string \newlabel {#1}{{#2}{\thepage}{#2}{#1}{}} }%
   \hypertarget{#1}{#2}
}
\makeatother

\newcommand{\R}{\mathbb R}

\newcommand{\DD}{\mathbb D}

\newcommand{\N}{\mathbb N}

\newcommand{\CA}{\mathcal A}
\newcommand{\CB}{\mathcal B}
\newcommand{\CH}{\mathcal H}

\newcommand{\CP}{\mathcal P}

\newcommand{\CX}{\mathcal X}
\newcommand{\CY}{\mathcal Y}
\newcommand{\CJ}{\mathcal J}

\newcommand{\Strain}{S_{\mathrm{train}}}
\newcommand{\xtest}{x_{\mathrm{test}}}

\newcommand{\CD}{\mathcal D}

\newcommand{\st}{\ : \ }

\DeclareMathOperator*{\argmin}{argmin}

\DeclareMathOperator*{\E}{\mathbb E}
\let\P\relax
\DeclareMathOperator*{\P}{\mathbb P}

\DeclareMathOperator{\Exp}{Exp}
\DeclareMathOperator{\Trans}{Trans}
\DeclareMathOperator{\ag}{Ag}
\DeclareMathOperator{\PAC}{PAC}
\DeclareMathOperator{\outDeg}{outDeg}

\newcommand{\bp}{\mathrm{BP}}
\newcommand{\Hall}{\mathsf{Hall}}
\newcommand{\test}{\mathrm{test}}

\newcommand{\defn}[1]{\textbf{#1}}
\newcommand{\edefn}[1]{\emph{\textbf{#1}}}

\DeclareRobustCommand{\stirling}{\genfrac\{\}{0pt}{}}

\newtheorem{innercustomgeneric}{\customgenericname}
\providecommand{\customgenericname}{}
\newcommand{\newcustomtheorem}[2]{%
  \newenvironment{#1}[1]
  {%
   \renewcommand\customgenericname{#2}%
   \renewcommand\theinnercustomgeneric{##1}%
   \innercustomgeneric
  }
  {\endinnercustomgeneric}
}

\newcustomtheorem{customprop}{Proposition}
\newcustomtheorem{customtheorem}{Theorem}

\newcommand{\maxent}{\texttt{MaxEnt}}
\newcommand{\email}[1]{\textsf{#1}}

\newenvironment{proofsketch}{
  \proof}{\endproof}

\title{Regularization and Optimal Multiclass Learning}

\author{Julian Asilis \\ USC \\ \email{asilis@usc.edu} \and 
Siddartha Devic \\ USC \\ \email{devic@usc.edu} \and 
Shaddin Dughmi \\ USC \\ \email{shaddin@usc.edu} \and  
Vatsal Sharan \\ USC \\ \email{vsharan@usc.edu} \and 
Shang-Hua Teng \\ USC \\ \email{shanghua@usc.edu}
}

\date{}

\begin{document}
\maketitle

\begin{abstract}
The quintessential learning algorithm of empirical risk minimization (ERM) is known to fail in various settings for which uniform convergence does not characterize learning. Relatedly, the practice of machine learning is rife with considerably richer algorithmic techniques, perhaps the most notable of which is regularization. Nevertheless, no such technique or principle has broken away from the pack to characterize optimal learning in these more general settings. 

The purpose of this work is to precisely characterize the role of regularization in perhaps the simplest setting for which ERM fails: multiclass learning with arbitrary label sets. Using \emph{one-inclusion graphs (OIGs)}, we exhibit optimal learning algorithms that dovetail with tried-and-true algorithmic principles: \emph{Occam's Razor} as embodied by \emph{structural risk minimization (SRM)}, the \emph{principle of maximum entropy}, and \emph{Bayesian} inference. We also extract from OIGs a combinatorial sequence we term the \emph{Hall complexity}, which is the first to characterize a problem's \emph{transductive error rate} exactly. 

Lastly, we introduce a generalization of OIGs and the \emph{transductive learning} setting to the agnostic case, where we show that optimal orientations of Hamming graphs --- judged using nodes' outdegrees minus a system of node-dependent \emph{credits} --- characterize optimal learners exactly. We demonstrate that an agnostic version of the Hall complexity again characterizes error rates exactly, and exhibit an optimal learner using maximum entropy programs. 
\end{abstract}

\newpage 

\begingroup
  \hypersetup{hidelinks}
  \tableofcontents
\endgroup

\newpage

\section{Introduction}\label{sec:intro}

The purpose of a machine learning algorithm is to generalize sample information into an effective model for future prediction. The poster-child of learning algorithms, empirical risk minimization (ERM), proceeds by simply selecting an element of the underlying hypothesis class with best fit to the training data. Despite its success over an impressive array of learning problems, ERM is known to fail catastrophically for many learnable problems, including even mild generalizations of binary classification \citep{shalev2010learnability, alon2022theory}.

Relatedly, machine learning has a rich algorithmic history of formulating learning as an optimization problem with a more carefully chosen objective function expressing the interplay between sample performance and generalization, bias and variance, or various other learning desiderata. In particular, a heavily used algorithmic principle is that of \emph{regularization}, the most familiar and explicit form of which is \emph{structural risk minimization} (SRM).
SRM adds a regularization term to the empirical risk when defining the objective to be minimized over the underlying hypothesis class. Informally, the regularizer is often meant to encode some notion of {\em hypothesis complexity}, such that the combined objective function can be viewed as an implementation of the principle of \emph{Occam’s Razor}.
The practice of machine learning is chock full of such regularization techniques for controlling model capacity, with impressive scientific and societal impact. Nonetheless, no such algorithmic technique or principle has broken away from the pack to characterize optimal learning in these more general settings. In particular, it is not known whether SRM or any particular regularization technique is sufficiently powerful to characterize optimal learning. It is this gap between theory and practice which motivates our work.

The purpose of this paper is to characterize the power of regularization in possibly the simplest setting for which ERM fails: multiclass learning.
We are inspired by the result of \citet{DS14} that there are learnable problems in multiclass learning that are not learnable by any proper learner (i.e., a learner that always emits an element of the underlying hypothesis class).
This impossibility result has direct ramifications to the framework of structural risk minimization. In particular, it implies that any learner witnessed as an optimization problem over the underlying hypothesis class $\CH$ is obligated to fail on some multiclass problems. Notably, this includes the standard toolkit of SRM. 

This motivates us to address the following fundamental question:

\begin{quote}
{\em What is the minimal augmentation to classical SRM that allows it to learn all (learnable) \linebreak multiclass problems?}
\end{quote}

Our first augmentation, necessary in order to bypass the properness obstruction discovered by \linebreak \citet{DS14}, is to grant regularizers access to the test point as input. Formally, such a \emph{local regularizer} is a map $\psi: \CH \times \CX \to \R_{\geq 0}$ for $\CX$ the domain set. We demonstrate, however, that local regularizers remain insufficiently expressive to learn certain multiclass problems. We proceed to consider regularizers that furthermore receive as input the sequence of unlabeled datapoints in the training set. We call these regularizers \emph{local unsupervised regularizers} --- formally, they are defined as maps $\psi: \CH \times \CX^n \times \CX \to \R_{\geq 0}$ for training sets of size $n$. Intuitively, the \textit{local} and \textit{unsupervised} modifications to the regularizer allow its induced learner to (1) be improper; and (2) perform an unsupervised pre-training step in which it uses the unlabeled sample data in order to establish local preferences over hypotheses in $\CH$. 

\paragraph{Contributions.}
As our primary technical contribution, we demonstrate that local unsupervised regularizers characterize multiclass learnability and, in fact, optimal learning.
That is, a multiclass problem is learnable if and only if there exists a local unsupervised regularizer whose SRM learners all learn the problem. 
Furthermore, in this event there is guaranteed to exist an unsupervised regularizer whose SRM learners are all optimal.
Importantly, we also demonstrate that our modification of SRM is \emph{minimal}: disallowing the regularizer access to either the test point or the unlabeled training set leads to the existence of learnable problems which cannot be learned by SRM using either of these weaker regularizers. 

\paragraph{Techniques and broad connections.}
Our characterization of local unsupervised regularization is enabled by the beautiful and insightful \emph{one-inclusion graphs} (OIGs), which have been used to model classification in the transductive setting starting with the work of \cite{haussler1994predicting}.
Through Hall's theorem and its generalization to infinite graphs, we derive from OIGs a complexity measure which exactly characterizes the optimal transductive error of a classification problem --- we aptly name this the \emph{Hall complexity}. 
We then use OIGs to derive learners which follow tried-and-true algorithmic principles: \emph{Occam's razor} as embodied by structural risk minimization (SRM), the \emph{principle of maximum entropy}, and \emph{Bayesian} inference. 

We provide two instantiations of an optimal algorithm within the local unsupervised SRM framework. 
The first is a deterministic learner implicit in the work of \cite{DS14}, whose regularization function we show to exist abstractly. We derive the second optimal learner, which is randomized, as the dual of a maximum-entropy convex program for orienting the OIG.
Its regularization function is the \emph{relative entropy} to a prior over hypotheses obtained through unsupervised learning. In addition to validating the maximum entropy principle in learning, this second algorithm can also be interpreted as a Bayesian learner which samples from a posterior distribution over labels, relative to a prior derived through unsupervised learning.

Most of our contributions extend from the realizable to the agnostic setting, including the characterization of optimal transductive errors through the (agnostic) Hall complexity and the design of an optimal Bayesian learner. To enable this extension, we adapt OIGs to the agnostic case so as to characterize agnostic multiclass learning in the transductive model.

\subsection{Related Work}
\label{sec:related}

We give a brief overview of related work and defer a full discussion to Appendix~\ref{appx:related_work}.
The importance of multiclass learners beyond the classical paradigm of empirical risk minimization (ERM) was discussed by \citet{shalev2010learnability}, who exhibited a learnable class that is not learnable by any ERM learner. 
The history of one-inclusion graphs in learning theory commences with the seminal work of \citet{haussler1994predicting}, who employed the transductive learning setting and the OIG algorithm proposed by \cite{alon1987partitioning} to obtain error guarantees for VC classes.
\citet{DS14} advanced the theory of multiclass learning on several fronts by demonstrating that proper learners fail on learnable hypothesis classes, improving the analysis of transductive error rates, and introducing the DS dimension.

More recently, \cite{brukhim2022characterization} used OIGs to prove that the DS dimension characterizes multiclass learnability, whereas the Natarajan dimension does not. 
OIGs also form a key ingredient in the study of learnability for \emph{partial} concept classes \citep{alon2022theory, kalavasis2022multiclass} as well as recent advances in PAC bounds for various problems \citep{aden2023optimal}, a characterization of learnability for realizable regression \citep{attias2023optimal}, and the design of optimal learners in the robust setting \citep{montasser2022adversarially}, to name only a few contributions.

Regarding regularization, perhaps most related to our theoretical formalization of regularizers is the work of \citet{hopkins2022realizable}, who consider the task of transforming realizable learners into agnostic learners.
In particular, the agnostic learners they produce can be seen as a type of unsupervised regularization, though not described as so in their work. Their learners use unlabeled sample data to restrict focus to a collection of hypotheses $F$ on which they perform ERM. When $F \subseteq \CH$, this restriction can be seen as a ``hard" regularizer assigning value $\infty$ to hypotheses in $F$ and zero otherwise. Note, however, that $F \subseteq \CH$ only if the original realizable learner is proper. (And, as we have seen, there exist learnable multiclass problems without any proper learners.) Furthermore, they use distinct datasets for regularization and risk minimization, while we do not. Lastly, their work begins with a realizable learner, whereas we are primarily concerned with the design of learners ``from scratch.''

\section{Preliminaries}\label{sec:prelims}

\subsection{Notation}
For $m \in \N$, we use $[m]$ to denote the set $\{1, \ldots, m\}$. For a predicate $P$, we let $[P]$ denote the Iverson bracket of $P$, i.e., 
$[P] = 1$ if $P$ is true and 0 if $P$ is false.
When $Z$ is a set, we use $Z^{< \omega}$ to denote the collection of all finite sequences in $Z$, i.e., $Z^{< \omega} = \bigcup_{i=1}^\infty Z^i$. When $Z$ is finite, we use $\Delta_Z$ to denote the set of all probability measures over $Z$, 
\[ \Delta_Z = \left \{P \colon Z \to \R_{\geq 0} : \sum_{Z} P(z) = 1 \right \}.\] 
For a tuple $S = (z_1, \ldots, z_n)$, we let $S_{-i}$ denote $S$ with its $i$th entry removed, as in  
$S_{-i} = (z_1, \ldots, z_{i-1}, z_{i+1}, \ldots, z_n).$
For $x \in \R$, $\lfloor x \rfloor$ denotes the greatest integer weakly less than $x$.

\subsection{Learning Theory}

We first recall the standard toolkit of supervised learning. A learning problem is determined by a \defn{domain} $\CX$, a \defn{label set} $\CY$, and a \defn{hypothesis class} $\CH \subseteq \CY^\CX$. We will refer to any function $\CX \to \CY$ as a \defn{hypothesis} or \defn{predictor}. Learning also requires a \defn{loss function} $\ell : \CY \times \CY \to \R$ to quantify a predictor's quality, perhaps the most fundamental of which is the 0-1 loss: $\ell_{0-1}(y, y') = [y \neq y'].$

We refer to learning problems with the 0-1 loss as \defn{multiclass classification} problems when $|\CY| > 2$ and \defn{binary classification} problems when $|\CY| = 2$. A \defn{labeled datapoint} is a pair $(x, y) \in \CX \times \CY$, and an \defn{unlabeled datapoint} is an element of $\CX$. We will occasionally refer to an (un)labeled datapoint merely as a datapoint when clear from context. A \defn{training set} $S$ is a finite tuple of labeled datapoints, i.e., $S \in (\CX \times \CY)^{< \omega}$. We may refer to them as \emph{training samples} or simply \emph{samples}. 

A \defn{learner} $A$ is a (possibly randomized) function from training sets to hypotheses, i.e.,  $A: (\CX \times \CY)^{< \omega} \to \CY^\CX$.  The \defn{true error}, or simply \defn{error}, incurred by a hypothesis $h \in \CY^\CX$ with respect to a distribution $D$ over $\CX \times \CY$ is the average loss its predictions incur on labeled datapoints drawn from $D$, i.e., $L_D(h) = \E_{(x, y) \sim D}  \ell(h(x), y)$. 
The \defn{empirical error} incurred by a hypothesis $h$ on a sample $S = \big((x_1, y_1), \ldots, (x_n, y_n)\big)$ is the average loss it suffers over datapoints in $S$, as in 
\[ L_S(h) = \frac{1}{n} \sum_{i=1}^n \ell(h(x_i), y_i). \] 

\begin{definition}
A learner $A$ is an \edefn{empirical risk minimizer} (ERM) with respect to $\CH$ if for all samples $S$, we have that $A(S) \in \argmin_\CH L_S(h).$
\end{definition}

A related technique is structural risk minimization (SRM), which amounts to empirical risk minimization along with an inductive bias favoring certain hypotheses in $\CH$ over others. 
This bias is formalized by a complexity measure known as a \emph{regularizer}.

\begin{definition}
A \edefn{regularizer} for a hypothesis class $\CH$ is a function $\psi : \CH \to \R_{\geq 0}$. 
A learner $A$ is a \edefn{structural risk minimizer} (SRM) with respect to $\CH$ if there exists a regularizer $\psi$ for $\CH$ such that, for all samples $S$, we have that 
\[ A(S) \in \argmin_{\CH} L_S(h) + \psi(h). \] 
\end{definition}

A learning problem is also defined by a criterion for success, the most celebrated of which is certainly Valiant's PAC learning framework \citep{valiant1984theory}. 

\begin{definition}\label{def:PAC_learning}
Let $\DD$ be a collection of probability measures over $\CX \times \CY$ and $\CH \subseteq \CY^\CX$ a hypothesis class. A learner $A$ is a \edefn{PAC learner for $\CH$ with respect to $\DD$} if there exists a \edefn{sample function} $m : (0, 1)^2 \to \N$ such that the following condition holds: for any $D \in \DD$ and $\epsilon, \delta \in (0, 1)$, a $D$-i.i.d. sample $S$ with $|S| \geq m(\epsilon, \delta)$ is such that, with probability at least $1 - \delta$ over the choice of $S$ and any internal randomness in $A$, 
\[ L_D(A(S)) \leq \inf_\CH L_D(h) + \epsilon. \] 
\end{definition}

In Definition~\ref{def:PAC_learning}, when $\DD$ consists of all probability measures over $\CX \times \CY$, one says that $A$ is an \defn{agnostic PAC learner} for $\CH$. When $\DD$ consists of all probability measure $D$ such that $L_D(h) = 0$ for some $h \in \CH$, one says that $A$ is a \defn{realizable PAC learner} for $\CH$. Hereafter, we will suppress dependence on the family $\DD$ and trust that it be established clearly in the surrounding context. 

\begin{definition}
The \edefn{sample complexity} of a learner $A$ with respect to a hypothesis class $\CH$,  $m_{\PAC, A} : (0, 1)^2 \to \N$, is the minimal sample function it attains as a learner for $\CH$. The  \edefn{sample complexity} of a class $\CH$ is the pointwise minimal sample complexity attained by any of its learners, i.e., 
$m_{\PAC, \CH}(\epsilon, \delta) = \min_A m_{\PAC, A}(\epsilon, \delta).$
\end{definition}

It will be useful to note that ERM and SRM learners are \emph{proper} learners, as we now define. 

\begin{definition}
A learner $A$ is said to be \edefn{proper with respect to $\CH$} if the guarantee $A(S) \in \CH$ holds for all samples $S$. 
\end{definition}

We remark briefly that we take a rather hands-off approach to measure-theoretic details throughout, adopting standard assumptions from e.g.\ \cite{shalev2014understanding}. See, e.g., \cite{BEHW89}, \cite[Appendix C]{pollard2012convergence} for a more precise treatment.

\subsection{Error: High-probability, Expected, and Transductive}
\label{sec:error}

The PAC learning framework demands high-probability guarantees of its learners, as presented in Definition~\ref{def:PAC_learning}. Though certainly the most standard framework for assessing learners, it is by no means the only one. Perhaps the most straightforward measure of a learner's quality is simply its expected error. This is the \emph{prediction model} of learning proposed by \cite{haussler1994predicting}.  Here, we require learners to attain vanishingly small expected error when trained on increasingly large samples, and endow such learners with corresponding sample complexities $m_{\Exp}(\epsilon)$. A third notion of error is that of the transductive error, to be described precisely shortly, which captures a learner's performance in the \emph{transductive model} of learning. This is a somewhat more adversarial version of the prediction model, and is in fact also described and used by \cite{haussler1994predicting}.\footnote{See also prior discussion of transductive learning by \cite{vapnik1974theory} and \cite{vapnik1982estimation}.} Once again, this framework requires vanishing transductive error of its learners, thereby equipping them with sample complexities $m_{\Trans}(\epsilon)$.

A natural question to ask of these learning criteria is as follows: How do the sample complexities they induce on classes $\CH$ differ? Does there exist a class $\CH$ whose sample complexities $m_{\Exp}(\epsilon)$ and $m_{\PAC}(\epsilon,\delta)$ scale considerably differently with $\epsilon$ (or $\delta$) than its transductive sample complexity $m_{\Trans}(\epsilon)$? For the case of realizable learning with bounded loss, the three frameworks turn out to be essentially equivalent. Notably, this places our study of one-inclusion graphs --- which are tailored to minimizing transductive error --- on firmer theoretical footing.

\begin{definition}\label{Definition:trans-learning-error-main-body}
The \edefn{transductive learning} setting is that in which the following steps take place: 
\begin{enumerate}
  \setlength\itemsep{0 em}
  \item An adversary chooses a collection of $n$ unlabeled datapoints $S = (x_1, \ldots, x_n) \in \CX^n$, along with a hypothesis $h \in \CH$. 
  \item The unlabeled data $S$ is revealed to the learner. 
  \item One datapoint $x_i$ is selected uniformly at random from $S$. The remaining datapoints $S_{-i}$ and their labels under $h$ are displayed to the learner. That is, the learner receives $(x_j, h(x_j))_{j \neq i}$. 
  \item The learner is prompted to predict the label of $x_i$, i.e., $h(x_i)$. 
\end{enumerate}
We refer to $x_i$ as the \edefn{test datapoint}, and the remaining $S_{-i}$ as the \edefn{training datapoints}. The \edefn{transductive error} incurred by a learner $A$ on the instance $(S, h)$ is its expected error over the uniformly random choice of $x_i$. That is, 
\[ L_{S, h}^{\Trans} (A) = \frac{1}{n} \sum_{i \in [n]} \ell \big( A(S_{-i}, h)(x_i), h(x_i) \big), \] 
where $A(S_{-i}, h)$ denotes the output of $A$ on the sample of datapoints in $S_{-i}$ labeled by $h$. 
\end{definition}

Intuitively, transductive error can be thought of as a fine-grained form of expected error that demands favorable performance on each individual sample $S$, and that furthermore ``hard-codes'' a uniform distribution over the datapoints of $S$. In particular, note the lack of an underlying distribution $D$ in the transductive setting. 

The \defn{transductive error rate} of a learner $A$ or class $\CH$ is defined respectively as 
\[ \epsilon_{A, \CH}(n) = \max_{S \in \CX^n, h \in \CH} L_{S, h}^{\Trans}(A), \; \text{  and  } \; \epsilon_{\CH}(n) = \min_{A} \epsilon_{A, \CH}(n). \] 

We show that the sample complexities of learning in the PAC, expected, and transductive settings differ by at most logarithmic factors in the realizable case. We emphasize that the content of this claim is neither novel nor particularly profound. Nevertheless, we believe that the community may benefit from a singular, organized treatment of the topic, which --- to our knowledge --- does not at present appear in the literature. See Appendix \ref{Appendix:3-errors} for further detail, along with \cite{dughmi2024transductive}. 

\begin{proposition}[Informal Proposition \ref{Proposition:3_errors_equivalent}]\label{Proposition:informal_3_errors_equivalent}
Fix a hypothesis class $\CH \subseteq \CY^\CX$ and a loss function taking values in $[0, 1]$. Let $m_{\PAC, \CH}$, $m_{\Exp, \CH}$, and $m_{\Trans, \CH}$ be the sample complexities of learning $\CH$ in the realizable PAC, expected, and transductive settings, respectively. Then the following inequalities hold for all $\epsilon, \delta \in (0, 1)$ and the constant $e \approx 2.718$.
\begin{enumerate}
  \item $m_{\Exp, \CH}(\epsilon + \delta) \leq m_{\PAC, \CH}(\epsilon, \delta) \leq O \left( m_{\Exp, \CH}(\epsilon / 2) \cdot \log (1 / \epsilon) \right)$. 
  \item $m_{\Exp, \CH}(\epsilon) \leq m_{\Trans, \CH}(\epsilon) \leq m_{\Exp, \CH}(\epsilon / e)$. 
\end{enumerate}
\end{proposition}
So far, we have restricted attention to realizable learning. We describe agnostic analogues of transductive learning and of Proposition~\ref{Proposition:informal_3_errors_equivalent} in Section~\ref{Section:agnostic-learning}.

\section{One-inclusion Graphs and the Hall Complexity}\label{sec:Hall}

One-inclusion graphs (OIGs) are powerful combinatorial objects that capture the structure of realizable learning under the 0-1 loss. They are particularly well-suited for analyzing transductive error, as defined in Definition~\ref{Definition:trans-learning-error-main-body}.
We begin the section by briefly reviewing OIGs, and refer the reader to Appendix~\ref{appendix:Hall} for a more complete (and self-contained) treatment. We then introduce the \emph{Hall complexity} derived from OIGs, which we show exactly characterizes a problem's optimal transductive error rate. Throughout the section, we remain in the setting of realizable multiclass classification. 

\begin{definition}\label{Definition:OIG_realizable}
Let $\CX$ be a domain, $\CY$ a label set, and $\CH \subseteq \CY^\CX$ a hypothesis class. The \edefn{one-inclusion graph} of $\CH$ with respect to $S \in \CX^n$, denoted $G(\CH|_S) = (V, E)$, is the following hypergraph: 
\begin{itemize}
    \setlength\itemsep{0 em}
    \item $V = \CH|_S$, and 
    \item $E = \bigcup_{i = 1}^n \CH|_{S_{-i}}$, where $e = h \in \CH|_{S_{-i}}$ is incident to all $g \in \CH|_S$ such that $g|_{S_{-i}} = h$. 
\end{itemize}
\end{definition}

Intuitively, each edge $e$ in $G(\CH|_S)$ corresponds to a labeled training sample and an unlabeled test point $\xtest$. A learner chooses a manner of completing $e$ into a fully labeled dataset by predicting a label for $\xtest$; this is precisely a choice of node incident to $e$ (i.e., an \emph{orientation} of $e$). In this view, the transductive error incurred on an instance $(S, h)$ equals the number of edges incident to $h|_S \in V$ which were \emph{not} oriented towards $h|_S$ --- that is, the number of times the learner ``should have" completed the label of $\xtest$ in accordance with the ground truth $h|_S$ but did not. This is simply the outdegree of the node $h|_S$. Let us formalize these concepts in the following definition.

\begin{definition}
\label{def:orientation}
An \edefn{orientation} of a hypergraph $G = (V, E)$ is a function $f: E \to V$ such that $f(e)$ is incident to $e$ for all $e \in E$. The \edefn{outdegree} of $v \in V$ in orientation $f$ is the number of edges $e$ incident to $v$ with $f(e) \neq v$. The \edefn{indegree} of $v$  in $f$ is the number of edges $e$ with $f(e)=v$. We say $G$ is \edefn{$\alpha$-orientable} if it can be oriented so that the in-degree of each vertex is at least $\alpha$.  Similarly, $G$ is \edefn{$\alpha$-coorientable} if it can be oriented so that the out-degree of each vertex is at most $\alpha$. We refer to orientations satisfying these conditions as \edefn{$\alpha$-orientations} and \edefn{$\alpha$-coorientations}, respectively. 
\end{definition}

We will also consider randomized orientations of OIGs, in which case we naturally extend Definition~\ref{def:orientation} so that a randomized $\alpha$-orientation is one which satisfies expected in-degree requirements, and likewise for coorientations. The following lemma formalizes the equivalence between learners attaining low transductive error and orientations of the OIG with low outdegree. 

\begin{lemma}\label{Lemma:trans_error_coorientations_main_text}
Let $\CA$ be a transductive learner for $\CH$. The following conditions are equivalent: 
\begin{enumerate}
\setlength\itemsep{0 em}
\item $\CA$ incurs transductive error at most $\epsilon$ on all samples of size $n$;
\item For each $h \in \CH$ and $S \in \CX^n$, $\CA$ induces an $(\epsilon \cdot n)$-coorientation on $G(\CH|_S)$; and  
\item For each $h \in \CH$ and $S \in \CX^n$, $\CA$ induces an $((1 - \epsilon) \cdot n)$-orientation on $G(\CH|_S)$. 
\end{enumerate}
\end{lemma}
\begin{proofsketch}
Conditions (2.) and (3.) are equivalent as $G(\CH|_S)$ is regular; each node has degree $|S| = n$. Conditions (1.) and (2.) are equivalent by our previous reasoning, i.e., the outdegree of a node $h \in G(\CH|_S)$ (with respect to the orientation induced by $\CA$) counts the number of errors made by $A$ when $h$ is the underlying transductive instance. See Lemma~\ref{Lemma:trans_error_coorientations} for the formal proof. 
\end{proofsketch}

We now define the \emph{Hall complexity}, which characterizes transductive error rates exactly. When $G = (V, E)$ is an undirected hypergraph and $U \subseteq V$, we let $E[U] \subseteq E$ denote the collection of edges with at least one incident node in $U$. 

\begin{definition}
The \emph{Hall density} of a graph $G = (V, E)$ is $\Hall(G) = \inf_{\substack{U \subseteq V, \\ |U| < \infty}} \frac{|E[U]|}{|U|}$. 
\end{definition}

\begin{definition}
\label{def:hall_complexity}
The \edefn{Hall complexity} of a hypothesis class $\CH$ is the function $\pi_{\CH} : \N \to \N$ defined   
\[ \pi_\CH(n) = \max_{S \in \CX^n} n - \Hall\big(G(\CH|_S)\big). \] 
\end{definition}

\begin{proposition}[Informal Proposition~\ref{Proposition:Hall_seq_main_result}]
\label{prop:informal_hall_seq_realizable}
The Hall complexity of a class $\CH$ exactly characterizes the optimal transductive error rate of learning $\CH$. That is, $\epsilon_\CH(n) = \frac{\pi_\CH(n)}{n}$ for all $n \in \N$.\footnote{We note briefly that the Hall complexity exactly characterizes the error rate of learning with arbitrary learners, which are permitted to use internal randomness. Introducing a floor in the definition of the Hall density begets a version of the Hall complexity which characterizes error rates with respect to deterministic learners.}
\end{proposition}

The proof is essentially an application of Hall's theorem to the (bipartite) edge-vertex incidence graph of the one-inclusion graph.
For a discussion of the Hall complexity's relation to other dimensions and sequences related to transductive error, we refer the reader to Remark~\ref{rmk:trans_error_charac_history}.

\section{Structural Risk Minimization}\label{sec:SRM}

In this section, we establish our main results in Theorems~\ref{Theorem:unsupervised-local-srm-factor2} and \ref{thm:realizable_randomized_learner}: realizable multiclass learnability is characterized by generalized regularizers granted access to the test point and unlabeled training points as input. That is, a multiclass problem is learnable precisely when it can be learned by such a regularizer, in which case one such regularizer is guaranteed to produce optimal learners. We term these regularizers \emph{local unsupervised regularizers}, and support our central theorem using results from Sections~\ref{Subsection:Impossibility},~\ref{Subsection:Deterministic_SRM_kcore}, and \ref{Subsection:Random_SRM_max_ent}, which establish both sufficiency and --- in a precise sense --- minimality of this generalized regularizer. Our proofs also help to illustrate the role of unsupervised methods in multiclass learning. Notably, our insights rely crucially on the machinery of one-inclusion graphs. 
Throughout the section, we remain in the setting of realizable classification. 

\subsection{Impossibility Results}\label{Subsection:Impossibility}

We begin by collecting impossibility results concerning learning techniques which learn all multiclass classification problems possible. 
The first result, due to \cite{DS14}, establishes that proper learners are in general insufficient for multiclass classification problems. 

\begin{proposition}[{\citet[Lemma~2]{DS14}}]\label{Proposition:proper_fails_multiclass}
There exists a PAC learnable hypothesis class $\CH \subseteq \CY^\CX$, for infinite label set $\CY$, that is not learnable by any proper learner. 
\end{proposition}

Notably, this eliminates the possibility that ERM or SRM learners are sufficiently expressive to learn all multiclass problems possible. 
To see why a proper learner may be insufficient, imagine a setting in which a class $\CH$ contains hypotheses whose ``complexities"  vary considerably over the domain $\CX$. For instance, $h_0 \in \CH$ acts ``simply'' on $X_{0} \subseteq \CX$ and with complexity on $X_1 \subseteq \CX$, while $h_1 \in \CH$ does the opposite. Then, intuitively, a regularizer $\psi$ would like to alternate between favoring the functions $h_0$ and $h_1$, depending upon the location being queried. See Figure~\ref{figure:local_regularizer}.

\begin{wrapfigure}{R}{0.47\textwidth}
\centering
\includegraphics[width=0.9\linewidth]{figures/local_regularizer.tikz}

\begin{flushleft}
{Figure~\customlabel{figure:local_regularizer}{1}: \emph{Hypotheses $h_0$ and $h_1$, depicted in red and yellow respectively. A  local regularizer may favor the simplicity of $h_0$ on test points drawn from the right region of the domain, and the simplicity of $h_1$ on test points drawn from the left region.}}
\end{flushleft}

\vspace{-5 em}
\end{wrapfigure}

This reasoning naturally gives rise to the notion of a \emph{local regularizer} that is non-uniform with respect to $\CX$.
\begin{definition}
\label{Definition:local-SRM}
A \edefn{local regularizer} for a hypothesis class $\CH$ is a function $\psi : \CH \times \CX \to \R_{\geq 0}$. 
A learner $A$ is a \edefn{local structural risk minimizer} with respect to $\CH$ if there exists a local regularizer $\psi$ for $\CH$ such that, for all samples $S$ and $x \in \CX$, 
\[ A(S)(x) \in \Big\{ h(x) : h \in \argmin_{\CH} L_S(h) + \psi(h, x) \Big\}. \] 
In this case, we say that $A$ is a learner \edefn{induced} by $\psi$. We say that $\psi$ \edefn{learns} the class $\CH$ if all learners it induces are PAC learners for $\CH$. 
\end{definition}

We note that local regularizers (or similar ideas) have been previously considered in computer vision
\citep{wolf2008local, prost2021learning} and in the learning theory community \citep{bottou1992local}.
Nonetheless, we demonstrate that this augmention is insufficient: even local regularizers are required to fail on some learnable classification problems.

\begin{proposition}\label{Proposition:local_regularizer_fails_multiclass}
There exists a PAC learnable class $\CH$ which no local regularizer can learn.
\end{proposition}

We defer the proof of Proposition~\ref{Proposition:local_regularizer_fails_multiclass} to Appendix~\ref{appendix:local_regularizer_fails}, and note that it employs the \emph{first Cantor class} of \linebreak \citet{DS14}.
Having already equipped regularizers with the information of the test point, there remains only one additional source of information with which to empower regularizers: the sample $S$ itself. If we were to grant local regularizers full access to $S$, it is straightforward to see that they could induce any learner, rendering the characterization meaningless (see Appendix~\ref{Appendix:regularizer_induces_any_learner}).
We should ask, then, what is the weakest summary statistic of $S$ which can be supplied to regularizers in order to increase their power? Perhaps the most simple is $|S|$, the cardinality of $S$. 
Equipping regularizers with this information is both a powerful tool in the practice of machine learning\footnote{There is evidence in both theory and practice which suggests that sample size plays a crucial role in calibrating regularizers; see e.g.\ the sample-size dependent regularizer from \cite{shalev2010learnability} or the SVM regularization in \cite{wainer2017EmpiricalEvaluation, shalev2008svm}.
} and, notably, would allow them to learn the class $\CH_\infty$ used in the proof of Proposition~\ref{Proposition:local_regularizer_fails_multiclass}.

\begin{definition}\label{Definition:local-size-based-regularizer}
A \edefn{local size-based regularizer} for a hypothesis class $\CH$ is a function \linebreak $\psi: \CH \times \N \times \CX \to \R_{\geq 0}.$
A learner $A$ is said to be \edefn{induced} by a local size-based regularizer $\psi$ if for all samples $S$ and datapoints $x \in \CX$, 
\[ A(S)(x) \in \left\{ h(x) : h \in \argmin_\CH L_S(h) + \psi(h, |S|, x) \right \}. \]
We say that $\psi$ \edefn{learns} the class $\CH$ if all learners it induces are PAC learners for $\CH$. 
\end{definition}

\noindent We now articulate a conjecture: local size-based regularizers are insufficient for classification.

\begin{conjecture}\label{Conjecture:size-based-regularizer-fails}
There exists a PAC learnable class $\CH$ that is not learned by any local size-based regularizer.
\end{conjecture}

In Appendix~\ref{Appendix:support-for-conjecture} we provide a learnable hypothesis class which we suspect cannot be learned by any local size-based regularizer, towards justifying the conjecture. Furthermore, we note that a negative resolution to the conjecture would somewhat undermine the structure of OIGs themselves. That is, there would exist learners for all multiclass problems which use strictly less information than OIGs (i.e., $|S|$ rather than all its unlabeled data). Given the volume of work on OIGs and their insights for learning, such an outcome could be considered surprising.

\vspace{-0.2 cm}

\subsection{Deterministic Learning with Acyclic Orientations}\label{Subsection:Deterministic_SRM_kcore}
 
Given the collection of impossibility results above and our suspicion about the insufficiency of size-based regularizers, we turn to providing a regularizer which utilizes not only the cardinality of $|S|$, but indeed the entire unlabeled sample set.
For a sample $S = ((x_i, y_i))_{i \in [n]}$, we let $S_{\CX}$ denote the sequence of unlabeled datapoints in $S$, i.e, $S_{\CX} = (x_i)_{i \in [n]}$. 

\begin{definition}\label{Definition:local-unsupervised-regularizer}
A \edefn{local unsupervised regularizer}, or simply \emph{unsupervised regularizer}, for a hypothesis class $\CH$ is a function $\psi : \CH \times \CX^{<\omega} \times \CX \to \R_{\geq 0}.$
A learner $A$ is a \edefn{local unsupervised structural risk minimizer} with respect to $\CH$ if there exists a local unsupervised regularizer $\psi$ such that the following guarantee holds for all samples $S$ and datapoints $x \in \CX$: 
\[ A(S)(x) \in \Big\{ h(x) : h \in \argmin_{\CH} L_S(h) + \psi(h, S_\CX, x) \Big\}. \] 
In this case, we say that $A$ is a learner \edefn{induced} by $\psi$. We say that $\psi$ \edefn{learns} the class $\CH$ if all learners it induces are PAC learners for $\CH$. 
\end{definition}

The central result of this section is that local unsupervised regularizers are indeed sufficiently expressive to optimally learn all multiclass problems with the 0-1 loss. We note that the proof is constructive when $\CY$ is finite and employs a compactness argument for infinite $\CY$. 

\begin{theorem}\label{Theorem:unsupervised-local-srm-factor2}
Let $\CY$ be a finite or countable label set and $\CH \subseteq \CY^\CX$ a hypothesis class. Then $\CH$ has an unsupervised local regularizer $\psi$ whose induced learners all attain optimal transductive error up to a constant factor of 2. 
\end{theorem}
\begin{proofsketch}
The full proof is given in Appendix~\ref{appx:proof-srm-factor-2}. First suppose that $\CY$ is finite. It suffices to demonstrate that for each $S \in \CX^n$ there exists an acyclic orientation of $G(\CH|_S)$ that is optimal to within a factor of 2. (From the acyclic orientation, one can topologically sort $G(\CH|_S)$ and define an unsupervised local regularizer that is decreasing on layers.) The acyclic orientation, implicit in the work of \citet{DS14}, arises from the following \emph{$k$-core} algorithm: repeatedly remove the vertex of lowest degree from $G(\CH|_S)$ and place it in the last layer of a topological ordering. The outdegree of any vertex in this ordering is precisely its degree in the undirected (sub)graph of $G(\CH|_S)$ before it was removed. This is bounded above by the maximum subgraph density of $G(\CH|_S)$, which in turns bounds the error of its best learner up to a factor of 2. The case of infinite $\CY$ requires a compactness argument. 
\end{proofsketch}

Note that local unsupervised regularizers are \emph{minimal} in the following sense: If we were to disallow the test point $x$ from a regularizer's input, its induced learners would be proper and thus fail on learnable problems by Proposition~\ref{Proposition:proper_fails_multiclass}. If we restricted access of the regularizer to $S_\CX$, its corresponding learners would again fail on learnable problems by Proposition~\ref{Proposition:local_regularizer_fails_multiclass}.

\subsection{Randomized Learning with Maximum Entropy Distributions}\label{Subsection:Random_SRM_max_ent}

The deterministic learner from the previous section serves as a tool for implementing principled tie-breaking between hypotheses attaining zero empirical risk. In this section, we propose a tie-breaking rule for \textit{randomized learners} based upon the maximum entropy principle.
This rule also has the property of being Bayesian in nature: it corresponds to learning a Bayesian prior using the unlabeled samples $S_\CX$ and test datapoint $\xtest$, which is then updated to a suitable posterior distribution over $\CH|_{S_\CX \cup \{\xtest\}}$ once the labeled data is revealed to the learner.
We derive this learner from the solution of a certain maximum-entropy convex program. Hereafter, we let $S_\CX^+$ denote the collection of unlabeled training datapoints alongside the test datapoint, i.e., $S_\CX^+ = S_\CX \cup \{\xtest\}$. 

We restrict attention to finite but arbitrarily large label sets $\CY$ in this section, as a consequence of certain measurability issues. However, none of our results are parameterized by the size of the underlying label set, suggesting that they may admit extension to infinite label spaces (perhaps via compactness arguments, as in \citep{asilis2024transductive,brukhim2022characterization, DS14}). 

\begin{theorem}
\label{thm:realizable_randomized_learner}
There exists an optimal randomized learner which can be summarized as follows.
    \begin{enumerate}
        \setlength\itemsep{0 em}
        \item Upon receiving the unlabeled datapoints $S_\CX^+=(x_1,\ldots,x_n)$, including the test point, use a convex program to compute an optimal randomized orientation of $G(\CH|_{S_\CX^+})$ with maximum entropy, then derive a prior distribution $\rho$ over $\CH|_{S_\CX^+}$ by normalizing the dual variables.
        \item Given the index $i$ of the test point, and labels $y_j$ for all datapoints $x_{j \neq i}$, apply a Bayes update to $\rho$ in order obtain a posterior $\rho'$. This posterior corresponds to restricting the prior to hypotheses consistent with the provided labels, and rescaling accordingly.
        \item Sample a hypothesis $h$ from $\rho'$ and output $h(x_i)$ as the predicted label of $x_i$.
    \end{enumerate}
\end{theorem}
\begin{proofsketch}
    The full proof is given in Appendix~\ref{appx:proof_max_entropy}.
    First, we draw an equivalence between the OIG $G(\CH|_{S_\CX^+})$ and a certain bipartite representation of the OIG, $G_{\bp}$.
    The graph $G_{\bp}$ has the property that assignments from its left-hand side are equivalent to orientations in $G(\CH|_{S_\CX^+})$.
    We define a convex program akin to \citet{singh2014entropy} which (randomly) assigns the left-hand side of $G_{\bp}$ so as to guarantee optimal error while maximizing entropy subject to this constraint.
    Due to the fact that dual variables enforce local decisions in convex matching programs, we are able to back out a Bayesian update step from the learner while preserving optimal error guarantees.
\end{proofsketch}

By Theorem~\ref{thm:realizable_randomized_learner}, we have that when the true labels for the  training datapoints are revealed, the randomized optimal (maximum entropy) learner is Bayesian in that it updates its posterior over hypotheses to have minimum relative entropy to the prior $\rho$, constrained on outputting hypotheses consistent with the training data.
We argue that this is a local unsupervised SRM in a natural generalized sense. In particular, we expand the space of hypotheses to include distributions over hypotheses, and have our regularizer assign a complexity to each such randomized hypothesis. 

\begin{definition}
    Let $\psi: \Delta_\CH \times \CX^{<\omega} \times \CX \to \R_{\geq 0}$ be a local unsupervised regularizer for randomized hypotheses.\footnote{Technically, for any $(S_\CX, x_{\mathrm{test}}) \in \CX^{<\omega} \times \CX$, we need only consider distributions over $\CH|_{S_\CX^+}$ for $S_\CX^+ = S_\CX \cup \{\xtest\}$, i.e., elements of $\Delta_{\CH|_{S_\CX^+}}$. As $\CY$ is taken to be finite in this section, $\CH|_{S_\CX^+}$ will always be finite.} A (randomized) learner $A$ is said to be induced by the regularizer $\psi$ if for all samples $S$ and datapoints $x \in \CX$,
\[ 
    A(S)(x) = h(x) \text{ where } h \sim D \text{ for } D \in \argmin_{\Delta_\CH} \E_{h \sim D} \big[ L_S(h) \big] + \psi(D, S, x).
\]
\end{definition}

\begin{corollary}
\label{corr:randomized_entropy_regularizer}
    The learner from Theorem~\ref{thm:realizable_randomized_learner} can be realized as a local unsupervised regularizer for randomized hypotheses.
\end{corollary}

We defer the proof to Appendix~\ref{appx:proof_randomized_entropy_regularizer} and note that the regularizer of Corollay~\ref{corr:randomized_entropy_regularizer} depends upon the relative entropy (i.e., KL divergence) between the distribution $D$ and the Bayesian prior $\rho$. In addition to being an SRM in the generalized sense just described, our learner can also be interpreted as an instantiation of the maximum entropy principle. In particular, if the prior $\rho$ were uniform, then indeed our learner would sample from the maximum entropy distribution over hypotheses consistent with the training data. More generally, sampling from the distribution which hues most closely to the prior subject to the provided labels, as measured by relative entropy, is the natural generalization of the maximum entropy principle to incorporate prior knowledge. That is, our learner deviates as little as possible from the prior subject to consistency with the provided labels. \\

\noindent \textbf{Discussion.} A priori, the value of a randomized learner may be unclear given that the previous section derives a deterministic one.
The pioneering work of \citet{jaynes1957information} states the \textit{principle of maximum entropy}: one should choose prior probabilities consistent with available information so as to maximize the entropy of the system.
\citet{jaynes1957information} further states: ``it is \textit{unreasonable} to assign zero probability to any event unless our data really rules out the case.''
Coupled with the fact that the previous deterministic learners are highly contingent upon the choice of optimal orientation, it is natural to ask for a (randomized) learner making fewer arbitrary choices. Yet another advantage of the randomized learner from this section is that it generalizes to the agnostic setting, as we will see shortly, whereas the deterministic strategy of Section~\ref{Subsection:Deterministic_SRM_kcore} does not appear to.

\section{Agnostic Learning}\label{Section:agnostic-learning}
Our discussion of learning and one-inclusion graphs has thus far pertained to the realizable case. Indeed, the structure of one-inclusion graphs and the transductive learning setting depends crucially upon guarantees provided by the realizability assumption. 

We devote this section to the generalization of the one-inclusion graph and its accompanying insights to the agnostic case. In particular, we define an agnostic version of the OIG and demonstrate that our previous results holds for the agnostic OIG, with the exception of the deterministic learner from Theorem~\ref{Theorem:unsupervised-local-srm-factor2}. A more complete discussion of the agnostic OIG, including precise statements of results and their proofs, is deferred to Appendix~\ref{sec:generalizing_oig}.

\begin{definition}[Informal Definition~\ref{Definition:agnostic-trans-learning-setup}]
\edefn{Transductive learning in the agnostic case} is defined as in the realizable case (Definition~\ref{Definition:trans-learning-error-main-body}), except that the adversary can arbitrarily label their datapoints.
\end{definition}

To compensate for the increased difficulty of the agnostic case, and in accordance with the PAC definition of agnostic learning, an agnostic transductive learner $A$ is only judged relative to the best-in-class performance across $\CH$: \[ L_S^{\Trans}(A) = \frac 1n \sum_{i \in [n]} \ell(A(S_{-i})(x_i), y_i) - \inf_{h \in \CH} \frac 1n \sum_{i \in [n]} \ell(h(x_i), y_i). \]

Unfortunately, the sample complexities of transductive and PAC learning are not known to be as closely related in the agnostic case as they are in the realizable case. With a simple of simple use of Markov's inequality and a repetition argument, one can show that agnostic PAC sample complexities exceed their transductive counterparts by at most a factor of $1 / \epsilon$. This is, of course, not a lower order factor, but it may be a loose bound. See \cite{dughmi2024transductive} for further discussion of the relationship between agnostic PAC and transductive learning. 

\begin{definition}
Let $\CH \subseteq \CY^\CX$ be a hypothesis class.
The \edefn{agnostic one-inclusion graph} of $\CH$ with respect to $S \in \CX^n$, denoted $G_{\ag}(\CH|_S) = (V,E)$, is 
the hypergraph with: 
\begin{itemize}
    \item $V = \CY^n$, one node for each possible labeling of the $n$ datapoints, and
    \item $E = \bigcup_{i=1}^n \CY^n|_{S_{-i}}$, where $e \in \CY^n|_{S_{-i}}$ is incident to each $v \in \CY^n$ such that $v|_{S_{-i}} = e$.
\end{itemize}
\end{definition}
The agnostic OIG is sometimes referred to as the \emph{Hamming graph} \citep{brouwer2011spectra}. We note that \citet{long1998complexity} use a similarly expanded OIG to study binary classification under distribution shift, but to our knowledge no previous work has fully expanded upon the idea to analyze learning in the agnostic case (see Appendix~\ref{appx:related_work} for further discussion).

Analagously to the realizable case, agnostic learners are in close correspondence with orientations of the agnostic OIG (c.f.\ Lemma~\ref{Lemma:trans_error_coorientations_main_text}).
A crucial difference, however, is that each vertex $v$ in an agnostic OIG is endowed with a number of \textit{Hamming credits} reflecting its distance from the underlying class $\CH$. Informally, the Hamming credits are subtracted from the outdegree of $v$ in any orientation of the OIG, so that learners (equivalently, orientations) are judged only upon the ``excess outdegree" they induce on vertices. Semantically, the Hamming credits reflect the fact that learners for the agnostic case need only compete with the performance of the best hypothesis in $\CH$. 
In Definition~\ref{appx:agnostic_hall_complexity} we use Hamming credits to define a suitable generalization of the Hall complexity which retains the exact transductive error characterization of its realizable counterpart.

\begin{proposition}[Informal Proposition~\ref{Proposition:agnostic_Hall_seq_main_result}]
The agnostic Hall complexity of a class $\CH$ exactly characterizes the optimal agnostic transductive error rate of learning $\CH$.
\end{proposition}

Lastly, we demonstrate that the randomized learner of Section~\ref{Subsection:Random_SRM_max_ent} generalizes to the agnostic case. Informally, the convex program used to produce the maximum entropy learner in Theorem~\ref{thm:realizable_randomized_learner} applies to the agnostic case with nearly identical reasoning, as it is robust to the addition of nodes to the OIG and to non-uniform out-degree requirements (i.e., to the presence of Hamming credits). 

\begin{proposition}[Informal Proposition~\ref{prop:agnostic_dual_characterization}]
The Bayesian randomized learner from Theorem~\ref{thm:realizable_randomized_learner} and its associated randomized regularizer from Corollary~\ref{corr:randomized_entropy_regularizer} can be extended to the agnostic case.
\end{proposition}

\section{Conclusion}

In pursuit of an algorithmic template for multiclass classification, we study the role of regularization in multiclass learning. We first observe that classical regularizers $\psi: \CH \to \R_{\geq 0}$ are insufficient to learn multiclass problems owing to the work of \cite{DS14}, and extend this impossibility result to the more powerful \emph{local regularizers} which are given access to the test datapoint. We then consider \emph{unsupervised local regularizers}, which are regularizers granted access to both the unlabeled test datapoint and the collection of all unlabeled training points. By exploiting the connection between unsupervised local regularizers and acyclic orientations of one-inclusion graphs (OIGs), we provide deterministic transductive learners that are nearly optimal (up to a factor of 2) for all multiclass problems in the realizable case. We then demonstrate an optimal randomized transductive learner for both the realizable and agnostic settings by way of a certain maximum entropy program, and show it to be an unsupervised local SRM as well as a pre-trained Bayesian learner. As part of our efforts, we also generalize the one-inclusion graph to the agnostic case and define the \emph{Hall complexity} associated to a class $\CH$, which is the first to provide an exact combinatorial characterization of transductive error rates. Future work includes resolving our conjecture that local size-based regularizers, which are of intermediate power between local regularizers and unsupervised local regularizers, are insufficient to learn multiclass problems. It would also be of interest to design optimal (or nearly optimal) multiclass learners which are computationally efficient, and to study the role of regularization (local, unsupervised, and otherwise) beyond classification. 

\subsection*{Acknowledgements}

Julian Asilis was supported by the USC Viterbi School of Engineering Graduate School Fellowship and by the  Simons Foundation.
Siddartha Devic was supported by the Department of Defense through the National Defense Science \& Engineering
Graduate (NDSEG) Fellowship Program. 
Shaddin Dughmi was supported by NSF Grant  CCF-2009060.
Vatsal Sharan was
supported by NSF CAREER Award CCF-2239265 and an Amazon Research Award.
Shang-Hua Teng was supported in part by NSF Grant CCF-2308744 and the Simons Investigator Award from the Simons Foundation. We thank Li Han for discussions that inspired the questions considered in this work.
We thank Jim Ferry for pointing to the connection between Proposition~\ref{Proposition:3_errors_equivalent} and Stirling numbers of the second kind. We thank Yusuf Hakan Kalayci for finding a typo in the definition of the agnostic Hall complexity from a previous draft.
Any opinions, findings, conclusions, or recommendations expressed in this material are those of the authors and do
not necessarily reflect the views of any of the sponsors such as the
NSF.

\bibliographystyle{plainnat}
\bibliography{references}

\newpage 

\subsection*{Appendix Organization}

Appendices~\ref{Appendix:3-errors} and \ref{appendix:Hall} are self-contained sections expanding upon  claims already mentioned in the main body: the equivalence between the PAC, transductive, and expected learning frameworks (Appendix~\ref{Appendix:3-errors}), and the Hall complexity of one-inclusion graphs (Appendix~\ref{appendix:Hall}). Appendix~\ref{Appendix:srm} briefly expands upon two points mentioned in Section~\ref{Subsection:Impossibility}: the triviality of local regularizers permitted access to all of $S$, and our conjecture concerning the insufficiency of local size-based regularizers. 

Appendix~\ref{appx:omitted_proofs} is devoted to proofs which were omitted from the main text. Appendix~\ref{sec:generalizing_oig} extends the machinery of one-inclusion graphs and Hall complexity to the agnostic case in a self-contained manner, as advertised briefly in Section~\ref{Section:agnostic-learning}. We conclude in Appendix~\ref{appx:related_work} with an expanded discussion of related work.

\begin{appendices}
\section{Equivalence of Errors}\label{Appendix:3-errors}

The PAC learning framework demands high-probability guarantees of its learners, as presented in Definition~\ref{def:PAC_learning}. Though certainly the most standard framework for assessing learners, it is by no means the only one. Perhaps the most straightforward measure of a learner's quality is simply its expected error. This is the  \emph{prediction model} of learning proposed by \cite{haussler1994predicting}.  Here, we require learners to attain vanishingly small expected errors when trained on increasingly large samples, and endow such learners with corresponding sample complexities $m_{\Exp}(\epsilon)$. A third notion of error is that of the transductive error, to be described precisely shortly, which captures a learner's performance in the \emph{transductive model} of learning. This is a somewhat more adversarial version of the prediction model, and is in fact also described and used by \cite{haussler1994predicting}. Once again, this framework requires vanishing transductive error of its learners, thereby equipping them with sample complexities $m_{\Trans}(\epsilon)$.

The most natural question to ask of these criteria for learning is whether they determine the same collection of learnable classes. That is, does there exist any hypothesis class $\CH$ learnable under one such framework but not another? In the case of realizable learning with a bounded loss function --- as we primarily consider --- it can be shown with little difficulty that the frameworks coincide in this sense. A central concern of learning theory, however, is not merely whether a given hypothesis class $\CH$ can be learned, but moreover the sample complexity with which it can be learned. Consequently, one should ask of the three learning criteria: How do the sample complexities they induce on classes $\CH$ differ? Does there exist a class $\CH$ whose sample complexities $m_{\Exp}(\epsilon)$ and $m_{\PAC}(\epsilon,\epsilon)$ scale considerably differently with $\epsilon$ than its transductive sample complexity $m_{\Trans}(\epsilon)$? For a given learning problem, how do guarantees at the level of one error correspond to guarantees for the others, if at all? 

The purpose of this section is to review these concepts and study the conditions under which one can favorably transform a learner with guarantees in one such error regime into a learner with guarantees in another (i.e., with only a modest effect on sample complexity). This is a topic which has received relatively little attention from the learning theory community, and which we believe would benefit from a clear collection of existing results. Furthermore, it will place our study of one-inclusion graphs --- which are tailored to minimizing transductive error --- on firmer theoretical footing. In particular, we will show for the case of realizable learning with bounded loss that the three learning frameworks are essentially equivalent, by providing modest bounds (at most logarithmic) on the extent to which their sample complexities may differ.

\subsection{A Simple Equivalence}

Throughout the section, we fix an arbitrary domain $\CX$, label set $\CY$, hypothesis class $\CH \subseteq \CY^\CX$, and a bounded loss function $\ell$, which we normalize to take values in $[0, 1]$. We also direct our attention to learning in the realizable case, and let $\DD_\CH$ denote the collection of all $\CH$-realizable distributions over $\CX \times \CY$. That is, $\DD_\CH$ consists of those distributions $D$ for which some $h \in \CH$ satisfies $L_D(h) = 0$. 

\begin{definition}\label{def:m_expA}
The \emph{sample complexity} $m_{\Exp, A}: (0, 1) \to \N$ of a learner $A$ in the expected error framework is the function mapping $\epsilon$ to the minimal $m$ for which the following condition holds: 
\[ (\forall D \in \DD_\CH)(\forall m' \geq m) \E_{S \sim D^{m'}} L_D\big(A(S)\big) \leq \epsilon.  \]
\end{definition}

That is, $m_{\Exp, A}$ tracks the minimal number of samples required by $A$ to attain a desired level of expected error, with respect to any $D \in \DD_\CH$. Definition~\ref{def:m_expA} forms an appropriate definition at the level of individual learners, but our interest ultimately lies in proving claims at the level of hypothesis classes. That is, we require a notion of sample complexity for a class $\CH$. 

\begin{definition}
The \emph{sample complexity} of a hypothesis class $\CH$ in the expected error framework is the optimal sample complexity attained by its learners, i.e., 
\[ m_{\Exp, \CH}(\epsilon) = \min_A m_{\Exp, A}(\epsilon), \] 
where $A$ ranges over all learners. 
\end{definition}

\begin{definition}\label{Definition:trans-learning-error}
The \edefn{transductive learning} setting is that in which the following steps take place: 
\begin{enumerate}
  \setlength\itemsep{0 em}
  \item An adversary chooses a collection of $n$ unlabeled datapoints $S = (x_1, \ldots, x_n) \in \CX^n$, along with a hypothesis $h \in \CH$.
  \item The datapoints $S$ are displayed to the learner. 
  \item One datapoint $x_i$ is selected uniformly at random from $S$. The remaining datapoints 
  \[ S_{-i} = (x_1, \ldots, x_{i-1}, x_{i+1}, \ldots, x_n) \]
  and their labels under $h$ are displayed to the learner. That is, the learner receives the data of $\big(x_j, h(x_j)\big)_{x_j \in S_{-i}}$. 
  \item The learner is prompted to predict the label of $x_i$, i.e., $h(x_i)$. 
\end{enumerate}
We refer to $x_i$ as the \edefn{test datapoint}, and the remaining $S_{-i}$ as the \edefn{training datapoints}. The \edefn{transductive error} incurred by a learner $A$ on the instance $(S, h)$ is its expected error over the uniformly random choice of $x_i$. That is, 
\[ L_{S, h}^{\Trans} (A) = \frac{1}{n} \sum_{i \in [n]} \ell \big( A(S_{-i}, h)(x_i), h(x_i) \big), \] 
where we use $A(S_{-i}, h)$ to denote the output of $A$ on the sample consisting of datapoints in $S_{-i}$ labeled by $h$. The \edefn{transductive error rate} incurred by $A$ is the function $\epsilon_{A, \CH} : \N \to \R$ defined by 
\[ \epsilon_{A, \CH}(n) = \max_{S \in \CX^n, \; h \in \CH} L_{S, h}^{\Trans}(A). \] 
\end{definition}

Intuitively, transductive error can be thought of as a fine-grained form of expected error that demands favorable performance on each individual sample $S$, and that furthermore ``hard-codes'' a uniform distribution over the datapoints of $S$. In particular, note the lack of an underlying distribution $D$ in the transductive setting. 

\begin{definition}
The \emph{transductive sample complexity} $m_{\Trans, A} : (0, 1) \to \N$ of a learner $A$ is the function mapping $\delta$ to the minimal $m$ for which $\epsilon_{A, \CH}(m') < \delta$ for all $m' \geq m$. That is, 
\[ m_{\Trans, A}(\delta) = \min \{ m \in \N \st \epsilon_{A, \CH}(m') < \delta, \; \forall m' \geq m\}. \] 
\end{definition}

We are now equipped to present the central claim of the section: these sample complexities differ by at most logarithmic factors. We emphasize again that the content of the claim is neither novel nor particularly profound. Nevertheless, we believe that the community may benefit from a singular, organized treatment of the topic, which --- to our knowledge --- does not at present appear in the literature. 

\begin{proposition}\label{Proposition:3_errors_equivalent}
Fix an arbitrary domain $\CX$, label set $\CY$, and hypothesis class $\CH \subseteq \CY^\CX$. Use a loss function taking values in $[0, 1]$. Then the following inequalities hold for all $\epsilon, \delta \in (0, 1)$ and the constant $e \approx 2.718$.
\begin{enumerate}
  \item $m_{\Exp, \CH}(\epsilon + \delta) \leq m_{\PAC, \CH}(\epsilon, \delta) \leq O \left( m_{\Exp, \CH}(\epsilon / 2) \cdot \log (1 / \epsilon) \right)$. 
  \item $m_{\Exp, \CH}(\epsilon) \leq m_{\Trans, \CH}(\epsilon) \leq m_{\Exp, \CH}(\epsilon / e)$. 
\end{enumerate}
\end{proposition}
\begin{proof}
The first inequality of claim (1.) follows directly from the fact that the loss function is bounded above by 1. That is, any learner attaining error $\leq \epsilon$ with probability $\geq (1 - \delta)$ on sample of size $n$ automatically incurs an expected error of at most
\[ \epsilon \cdot ( 1 - \delta) +  1 \cdot \delta \leq \epsilon + \delta. \]
For the second inequality in (1.), let $A$ be a learner attaining expected error $\leq \epsilon / 2$ on samples of size $n$, from which we would like to extract a high-probability guarantee. The key observation is that $A$ can be boosted to attain expected error $\leq \epsilon$ with probability $\geq 1 - \epsilon$ using only an additional factor of $O(\log(1 / \epsilon))$ many examples. In particular, a sample of size $n \cdot O(\log(1/ \epsilon))$ can be divided into units of size $n$, half of which are used to train $A$ and produce candidate hypotheses $h_1, \ldots, h_\ell$, and half of which are used as a single validation set to select the best such $h_i$, as described in \cite{DS14}. 

The first inequality of claim (2.) follows from the standard leave-one-out argument of \cite{haussler1994predicting}. In particular, if $A$ is a learner incurring transductive error $\leq \epsilon$, then the same can be said of its expected error. For a sample $S = \big((x_1, y_1), \ldots, (x_n, y_n) \big)$, recall that $S_{-i}$ denotes the sample consisting of all labeled examples in $S$ other than $(x_i, y_i)$.
\begin{align*}
\E_{S \sim D^m} L_D\big(A(S)\big) &= \E_{\substack{S \sim D^m \\ (x, y) \sim D}} \ell \big(A(S)(x), y \big) \\ 
&= \E_{S \sim D^{m+1}} \ell \big(A(S_{-(m+1)})(x_{m+1}), y_{m+1} \big) \\
&= \E_{S \sim D^{m+1}} \E_{i \in_R [m+1]} \ell \big( A(S_{-i})(x_{i}), y_{i} \big) \\ 
&\leq \sup_{S} \E_{i \in_R [m+1]} \ell \big( A(S_{-i})(x_{i}), y_{i} \big) \\
&= \epsilon_{A, \CH}(m+1). 
\end{align*}
The second inequality in (2.) is claimed without proof by \cite{DS14}; in the interest of completeness, we provide a proof in Lemma~\ref{Lemma:c=e}. 
\end{proof}

We note briefly that Theorem 2.2 of \cite{aden2023optimal} provides tighter bounds than Proposition \ref{Proposition:3_errors_equivalent} for transforming a learner with optimal transductive error into one with PAC error guarantees. Notably, however, the theorem is phrased only for finite label sets, in contrast to the generality of Proposition \ref{Proposition:3_errors_equivalent}. The ability to quantify over arbitrary label sets is crucial for our purposes, as it is primarily over infinite label sets that the theory of multiclass classification departs from binary classification (e.g., uniform convergence fails to characterize learnability, ERM learners fail for learnable problems, etc.). As such, we are best served with the slightly looser but considerably more general statement of Proposition \ref{Proposition:3_errors_equivalent}.

\begin{remark}
It was recently shown in \cite{OIG-not-always-optimal} that one-inclusion graphs, which attain optimal transductive error, do not always provide optimal high-probability guarantees. We note that this is compatible with Proposition~\ref{Proposition:3_errors_equivalent}, which quantifies over all learners for a given class $\CH$ and does \emph{not} claim that a learner attaining optimal error in one regime need do so for the others as well. Proposition~\ref{Proposition:3_errors_equivalent} instead demonstrates that the levels of performance attained by the (possibly distinct) optimal learners for the three notions of error are comparable. This suffices to justify a focus on any of the three errors --- in our case, transductive --- as the sample complexities enjoyed by optimal learners in the other regimes will be comparable. (And in fact, the proof of Proposition~\ref{Proposition:3_errors_equivalent} provides a simple recipe for transforming optimal transductive learners into near-optimal high-probability learners.)
\end{remark}

\begin{lemma}\label{Lemma:c=e}
Fix an arbitrary domain $\CX$, label set $\CY$, and hypothesis class $\CH \subseteq \CY^\CX$. 
Then for all $\epsilon \in (0, 1)$, $m_{\Trans, \CH}(\epsilon) \leq m_{\Exp, \CH}(\epsilon / e)$. 
\end{lemma}
\begin{proof}
Fix $\epsilon$, set $n = m_{\Exp, \CH}(\epsilon)$, and let $A$ be a learner attaining expected error $\leq \epsilon$ on samples of size $n$. We will extract from $A$ a learning attaining transductive error at most $e \cdot \epsilon$, completing the proof. First, an intermediate result. 

\begin{quote}

\vspace{-0.6 cm}
\begin{lemma}\label{Lemma:coupon_collector}
For each $n \in \N$, there exists an $m_n \in \N$ such that $m_n$ independent draws from a uniform distribution over $n$ items results in seeing exactly $n - 1$ unique elements with probability at least $\frac 1e$. 
\end{lemma}
\begin{proof}
Let $\stirling ab$ denote Stirling numbers of the second kind. Then for $m$ many draws from the uniform distribution over $n$ elements, the probability of seeing exactly $n-i$ unique elements is 
\[ \frac{1}{n^m} \stirling{m}{n-i} \frac{n!}{i!}.  \]  
We refer to this event as $E_i$, when $n$ and $m$ are clear from context. For the moment let us shift our perspective so that $m$ is the variable of interest and our asymptotics are taken with respect to $m$. That is, for each given $m$, can we find an $n$ such that $m$ uniform draws result in exactly $n - 1$ unique elements with constant probability? 

Given $m$, let $k_m$ be a value maximizing $\stirling{m}{i}$ over $i \in [m]$. Then take $n_m = k_m + 1$. For each $i \in [m]$, we have that $P(E_i) \propto \stirling{m}{n_m-i} \cdot \frac{1}{i!}$. By definition of $n_m$, the first factor in this product is maximized for $i = 1$. Therefore, we can lower bound $P(E_i)$ by imagining instead that $P(E_i) \propto \frac{1}{i!}$. And in this case, clearly, $P(E_i) \geq \frac{1}{e}$. 

Lastly, by \cite{stirling}, each $n \in \N$ appears as $k_m$ for some $m$, completing the argument. 
\end{proof}
\end{quote}

We will now design a randomized learner $B$ that attains transductive error $\leq e \cdot \epsilon$ on samples of size $n$. In particular, $B$ acts as follows upon receiving sample $S = \big((x_1, y_1), \ldots, (x_{n-1}, y_{n-1}) \big)$ and test datapoint $x_{n}$: it randomly generates a uniform sample of size $m_n$ from the uniform distribution over $S$, call it $S'$, and returns $A(S')$.  

To analyze the transductive error of $B$, fix a transductive learning instance $(S, h)$. Let $S = (x_1, \ldots, x_n)$, and have $D$ denote the uniform distribution over $(x_1, h(x_1)), \ldots, (x_n, h(x_n))$. Call a sample drawn from $D$ \emph{good} if it contains exactly $n - 1$ unique elements. 
The crucial observation is as follows: the probability that $B$ errs, averaged over a uniformly random test point $x_i \in S$ and the randomness internal to $B$, equals precisely the probability that $A$ errs conditioned on receiving good samples of size $m_n$. By Lemma~\ref{Lemma:coupon_collector}, the latter quantity is at most $e \cdot \epsilon$. This completes the argument. 
\end{proof}

\section{One-inclusion Graphs and the Hall Complexity}\label{appendix:Hall}

One-inclusion graphs (OIGs) are powerful combinatorial objects that capture the structure of realizable learning under the 0-1 loss. They are particularly well-suited for analyzing transductive error, as defined in Definition~\ref{Definition:trans-learning-error}. In particular, it was demonstrated in \cite{DS14} that for a given hypothesis class $\CH$, a combinatorial sequence $\mu_\CH: \N \to \N$ associated to the one-inclusion graphs of $\CH$ provides a constant factor approximation to the optimal transductive error of its learners. The central result of this section is the introduction of a new sequence associated to the OIGs of $\CH$, which we term the \emph{Hall complexity}, that characterizes optimal transductive error exactly. To this end, we recall the appropriate definitions concerning one-inclusion graphs in Section \ref{subsec:OIG}, and present the Hall complexity in Section \ref{subsec:Hall}. 

Throughout the section we restrict focus to realizable learning under the 0-1 loss, over arbitrary domain and label sets $\CX, \CY$.

\subsection{One-inclusion Graphs (with a Bipartite Perspective)}\label{subsec:OIG}

Recall the basic structure of transductive learning: a learner is presented with $n$ unlabeled datapoints $S = ( x_1, \ldots, x_n )$, one such datapoint $i$ is removed uniformly at random from $S$, and the learner is asked to guess $h(i)$ from the data of $h|_{S_{-i}}$, for some $h \in \CH$. Finally, the learner is judged on its average performance over the randomness of $i \in [n]$, with respect to the 0-1 loss function. Note that, equipped with this loss function, the transductive error incurred by learner $A$ on the instance $(S, h)$ is 
\[  L_{S, h}^{\Trans} (A) = \frac{1}{n} \sum_{i \in [n]} \big[ A(S_{-i}, h)(x_i) \neq h(x_i) \big]. \] 

Now let us take the perspective of a transductive learner for the moment, and imagine that we have just been given the data of the $n$ datapoints $S = (x_1, \ldots, x_n)$, including the yet-to-be-selected test point. Our objective is to minimize the worst-case transductive error we incur over any possible ``ground truth'' $h \in \CH|_S$. First, a simple observation: upon observing $h|_{S_{-i}}$, we ought to output $g(x_i)$ for $g \in \CH|_S$ a function such that $g|_{S_{-i}} = h|_{S_{-i}}$. Otherwise, it is guaranteed that $g(x_i) \neq h(x_i)$ and thus that we incur a loss of 1, the maximal possible loss. That is, in realizable learning with the 0-1 loss, any sensible learner ought to be an ERM. Hereafter, we will only consider (transductive) learners obeying this mild property. 

We now introduce some notation. Let us represent each $g \in \CH|_S$ as a \emph{fully labeled dataset} 
\linebreak 
$\big((x_1, g(x_1)), \ldots, (x_n, g(x_n)) \big)$, i.e., by its graph. Similarly, we can represent each $h \in \CH|_{S_{-i}}$ as a \emph{partially labeled dataset} $\big( (x_1, h(x_1)), \ldots, (x_i, ?), \ldots, (x_n, h(x_n)) \big)$, i.e., by its graph augmented by a ``?'' accompanying the test datapoint $x_i$. For $g \in \CH|_S$ a fully labeled dataset and $h \in \CH|_{S_{-i}}$ a partially labeled dataset, we say that $g$ \emph{completes} $h$ (or is \emph{compatible} with $h$) when they agree on $S_{-i}$. 

In this light, by our previous reasoning, the task of a transductive learner is to complete each partially labeled dataset into a fully labeled dataset from among $\CH|_S$. Note also the following observations: 
\begin{enumerate}
    \item Each fully labeled dataset (i.e., ground truth $g \in \CH|_S$) is compatible with exactly $n$ partially labeled datasets, each corresponding to one location for the ``?''. 
    \item Upon making a choice of fully labeled dataset for each partially labeled dataset, the transductive error incurred on the ground truth $g \in \CH|_S$ is proportional to the number of compatible partially labeled datasets that are \emph{not} assigned to $g$. Equivalently, $n$ minus the number of partially labeled datasets that are assigned to $g$. 
\end{enumerate}

These insights are perhaps best expressed in the form of a bipartite graph $G_{\bp} = (\CA, \CB, E)$. Let the partially labeled datasets and fully labeled datasets form $\CA$ and $\CB$, respectively. We then have an edge $(u, v) \in E$ precisely when $v$ is a fully labeled dataset completing $u$, as depicted in Figure \ref{figure:bipartite}. The following claims follow immediately from our previous reasoning: a transductive learner on $S$ is precisely a choice of assignment in $G_{\bp}$ (mapping each $u \in \CA$ to an incident $v \in \CB$), and the worst-case transductive error it incurs over $h \in \CH|_S$ is determined by the minimal indegree of a node in $\CB$ under this assignment. 

In short, transductive learning devolves to finding assignments in bipartite graphs that maximize minimal indegrees in $\CB$. With only a slight change in perspective, we will arrive at one-inclusion graphs. Namely, given $G_{\bp} = (\CA, \CB, E)$, consider the hypergraph $G$ whose vertex set is $\CB$ and edge set is $\CA$, such that $a \in \CA$ is incident to precisely those vertices in $G$ (as an edge) with which it is incident in $G$ (as a node). Simply put, view each $a \in \CA$ as an edge rather than a node! 

This is formalized by the following definition. 

\begin{definition}
Let $\CX$ be a domain, $\CY$ a label set, and $\CH \subseteq \CY^\CX$ a hypothesis class. The \edefn{one-inclusion graph} of $\CH$ with respect to $S \in \CX^n$, denoted $G(\CH|_S)$, is the hypergraph defined by the following vertex and edge sets: 
\begin{itemize}
    \item $V = \CH|_S$, and 
    \item $E = \bigcup_{i = 1}^n \CH|_{S_{-i}}$, where $e = h \in \CH|_{S_{-i}}$ is incident to all $g \in \CH|_S$ such that $g|_{S_{-i}} = h$. 
\end{itemize}
We will sometimes write such an edge $e$ as $e_{(h, i)}$, with $h \in \CH|_{S}$ and $i \in [n]$. Under this representation, $e_{(h, i)}$ is incident to all nodes $g \in \CH|_S$ such that $g|_{S_{-i}} = h|_{S_{-i}}$. 
\end{definition}

Let us now formally define the bipartite view of one-inclusion graphs, $G_{\bp}$, which we have informally discussed. We will return frequently to this view of one-inclusion graphs throughout the work.

\begin{definition} \label{defn:bipartite_oig}
Let $\CX$ be a domain, $\CY$ a label set, and $\CH \subseteq \CY^\CX$ a hypothesis class. 
The \edefn{bipartite view of the one-inclusion graph} of $\CH$ with respect to $S \in \CX^n$, denoted $G_\bp(\CH|_S)$, is defined as follows. Let $G(\CH|_S) = (V, E)$ be the usual one-inclusion graph of $\CH$ with respect to $S$. Then $G_\bp(\CH|_S) = (L, R, E')$ is the bipartite graph in which $L = E$, $R = V$, and $(e, v) \in E'$ precisely when $e$ is incident to $v$ in $G(\CH|_S)$. In other words, $G_{\bp}(\CH|_S)$ is precisely the edge-vertex incidence graph of $G(\CH|_S)$. We may denote it simply by $G_\bp$ when $\CH$ and $S$ are clear from context. 
\end{definition}

\begin{remark}
There are various non-equivalent definitions of one-inclusion graphs in the literature. Ours is equivalent to that of \cite[Definition 9]{brukhim2022characterization}, and allows for hyperedges to have size 1 (i.e., to be self-loops), which crucially results in each node having degree exactly $n = |S|$. Originally, OIGs were defined by \cite{haussler1994predicting} with the requirement that edges have size at least 2 (see also \cite{alon1987partitioning}). For our purposes, this has the unfavorable consequence of permitting nodes to have different degrees, which prevents us from establishing an equivalence between maximizing nodes' indegrees and minimizing nodes' outdegrees (to be seen shortly). In addition to the usual graph-theoretic definitions of OIGs, it will be often be useful to retain the bipartite interpretation of Figure \ref{figure:bipartite} as a supplementary perspective. In particular, it will form the basis for analyzing the Hall complexity in a few moments, and for generalizing OIGs to the agnostic case in Appendix~\ref{sec:generalizing_oig}.
\end{remark}

Recall now the notion of an orientation of a hypergraph. 

\begin{definition}
An \edefn{orientation} of a hypergraph $G = (V, E)$ is a function $f: E \to V$ such that $f(e)$ is incident to $e$ for all $e \in E$. The \edefn{outdegree} of $v \in V$ in orientation $f$ is the number of edges $e$ incident to $v$ with $f(e) \neq v$. Similarly, the \edefn{indegree} of $v$  in $f$ is the number of edges $e$ with $f(e)=v$.
\end{definition}

The following is immediate from our previous reasoning. 

\begin{lemma}\label{Lemma:outdeg_is_trans_error}
There is a one-to-one correspondence between (deterministic) transductive learners for $\CH$ and orientations of $G(\CH|_S)$ for all $S \in \CX^{<\omega}$. Furthermore, a transductive learner $A$ incurs transductive error $\leq \epsilon$ on the instance $(h, S)$ if and only if $h|_S$ has outdegree $\leq \epsilon \cdot |S|$ in the graph $G(\CH|_S)$ oriented by $A$. 
\end{lemma}
\begin{proof}
Fix a learner $A$ and $S \in \CX^n$. The action of $A$ on partially labeled datasets induces an orientation on $G(\CH|_S)$, such that each hyperedge $e = \big( (x_1, y_1), \ldots, (x_i, ?), \ldots, (x_n, y_n) \big)$ is oriented toward $\big( (x_1, y_1), \ldots, (x_i, y_i), \ldots, (x_n, y_n) \big)$ for $y_i$ the output of $A$ on input $e$. We write $e \to h$ if $e$ is oriented towards $h$ and $e \not \to h$ otherwise. Now, for any $h \in \CH|_S$ we have: 
\begin{align*} 
L_{S, h}^{\Trans} (A) &= \frac{1}{n} \sum_{i \in [n]} \big[ A(S_{-i}, h)(x_i) \neq h(x_i) \big] \\
&= \frac 1n \sum_{i \in [n]} [e_{(h, i)} \not \to h ] \\
&= \frac 1n \cdot \outDeg(h).
\end{align*} 
\end{proof}

The previous lemma justifies a focus on orientations that minimize nodes' outdegrees. For one-inclusion graphs, in which every node has undirected degree exactly $|S|$, this amounts precisely to maximizing nodes' indegrees. 

\begin{definition}\label{Definition:alpha-(co)orientable}
Let $G$ be an undirected hypergraph. We say $G$ is \edefn{$\alpha$-orientable} if it can be oriented so that all its vertices' indegrees are at least $\alpha$. $G$ is \edefn{$\alpha$-coorientable} if it can be oriented so that all its vertices' outdegrees are at most $\alpha$. We will refer to orientations satisfying these conditions as \edefn{$\alpha$-orientations} and \edefn{$\alpha$-coorientations}, respectively. 
\end{definition} 

We may sometimes use \emph{randomized} orientations to describe randomized learners. In these cases, we naturally extend Definition~\ref{Definition:alpha-(co)orientable} so that a randomized $\alpha$-orientation is one which satisfies \emph{expected} in-degree requirements, and likewise for coorientations. 

\begin{lemma}\label{Lemma:trans_error_coorientations}
Let $A$ be a transductive learner for $\CH$. The following conditions are equivalent.
\begin{enumerate}
    \item $A$ incurs transductive error at most $\epsilon$ on all samples of size $n$.
    \item For each $h \in \CH$ and $S \in \CX^n$, $A$ induces an $(\epsilon \cdot n)$-coorientation on $G(\CH|_S)$.   
    \item For each $h \in \CH$ and $S \in \CX^n$, $A$ induces an $((1 - \epsilon) \cdot n)$-orientation on $G(\CH|_S)$. 
\end{enumerate}
\end{lemma}
\begin{proof}
Conditions (2.) and (3.) are patently equivalent, as any node in the undirected graph $G(\CH|_S)$ has degree exactly $n = |S|$, one for each point in $S$ that can be omitted. The equivalence between (1.) and (2.) follows immediately from Lemma~\ref{Lemma:outdeg_is_trans_error}. 
\end{proof}

\addtocounter{figure}{1}
\begin{figure}[t]
\centering
\definecolor{myblue}{RGB}{80,80,160}
\definecolor{mygreen}{RGB}{80,160,80}
\begin{tikzpicture}[thick,
  every node/.style={draw,circle},
  fsnode/.style={fill=myblue},
  ssnode/.style={fill=mygreen},
  every fit/.style={ellipse,draw,inner sep=-2pt,text width=2cm},
  ->,shorten >= 3pt,shorten <= 3pt
]

\begin{scope}[start chain=going below,node distance=7mm, minimum size=0.3cm]
  \node[fsnode,on chain] (f1) [label=left: {$(0, 0, ?)$}] [label={[label distance=0.3cm]above: {\scalebox{1.3}{$\CA$}}}] {};
  \node[fsnode,on chain] (f2) [label=left: {$(1, 0, ?)$}] {};
  \node[fsnode,on chain] (f3) [label=left: {$(0, 1, ?)$}] {};
  \node[fsnode,on chain] (f4) [label=left: {$(0, ?, 0)$}] {};
  \node[fsnode,on chain] (f5) [label=left: {$(1, ?, 0)$}] {};
  \node[fsnode,on chain] (f6) [label=left: {$(?, 0, 0)$}] {};
  \node[fsnode,on chain] (f7) [label=left: {$(?, 1, 0)$}] {};
\end{scope}

\begin{scope}[xshift=5 cm,yshift=-1.2cm,start chain=going below,node distance=12 mm]
  \node[ssnode,on chain] (s1) [label=right: {$(0, 0, 0)$}] [label={[label distance=0.3cm]above: {\scalebox{1.3}{$\CB$}}}]{};
  \node[ssnode,on chain] (s2) [label=right: {$(1, 0, 0)$}] {};
  \node[ssnode,on chain] (s3) [label=right: {$(0, 1, 0)$}] {};
\end{scope}

\draw (f1) -- (s1);
\draw (f2) -- (s2);
\draw (f3) -- (s3);
\draw (f4) -- (s1);
\draw (f4) -- (s3);
\draw (f5) -- (s2);
\draw (f6) -- (s1);
\draw (f6) -- (s2);
\draw (f7) -- (s3);
\end{tikzpicture}
\caption{Bipartite graph $G_\bp$ representing transductive learning on a traning set $S$ of three datapoints, for which $\CH$ can express the binary strings $(0, 0, 0)$, $(1, 0, 0)$, and $(0, 1, 0)$. We fix an ordering of the unlabeled data in $S$ and represent each fully or partially labeled dataset using only its labels. Note that each fully labeled dataset has degree exactly 3 = $|S|$. An optimal transductive learner amounts precisely to an assignment of $G$ (i.e., choice of incident edge for each vertex in $\CA$) that maximizes the minimal indegree in $\CB$. Here, the best indegree that can be attained is 2.}
\label{figure:bipartite}
\end{figure}
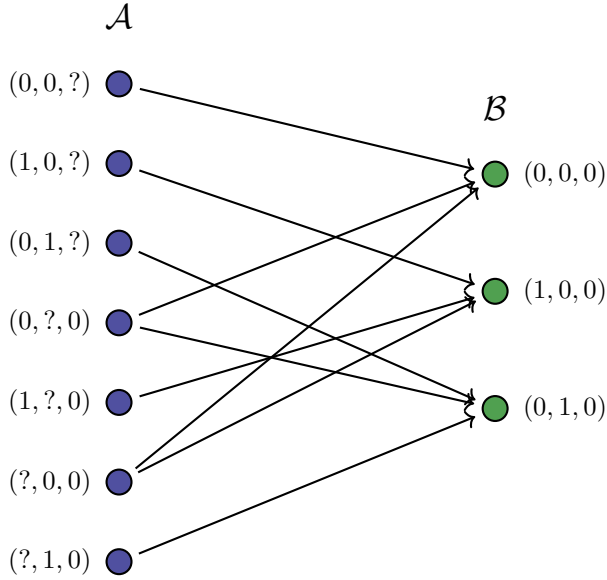

\subsection{The Hall Complexity}\label{subsec:Hall}

Given a framework for learning and a hypothesis class $\CH$, perhaps the most pressing question is: How quickly can $\CH$ be learned, if at all? For transductive learning, progress was first made on the issue by \citet{haussler1994predicting}, who demonstrated an upperbound on the transductive error rate of learning $\CH$ based upon the maximum subgraph density of its one-inclusion graphs. (See Remark~\ref{rmk:trans_error_charac_history} for further detail). Subsequently, \citet{DS14} introduced a sequence characterizing optimal transductive errors up to constant factors. We now introduce a combinatorial sequence that characterizes the errors of optimal transductive learners exactly. 

\begin{definition}
Let $G = (V, E)$ be an undirected hypergraph. For a set of nodes $U \subseteq V$, let $E[U] \subseteq E$ denote the collection of edges with at least one incident node in $U$. The \defn{Hall density} of $G$ is
\[ \Hall(G) = \inf_{\substack{U \subseteq V, \\ |U| < \infty}} \frac{|E[U]|}{|U|}. \]
\end{definition}

\begin{proposition}\label{Proposition:Hall_dimn_orientations}
Let $G$ be an undirected hypergraph in which each node has finite degree. Then $\Hall(G)$ is the supremum of all $\alpha$ for which $G$ is $\alpha$-orientable. 
\end{proposition}
\begin{proof}
An orientation of $G = (V, E)$ is precisely a (possibly randomized) assignment $E \to V$ in which each edge is assigned to an incident node.
In order for there to exist such an assignment in which each $v \in V$ receives $\alpha$ in-degree, it is clearly necessary that for each finite $U \subseteq V$, $|E[U]| \geq \alpha \cdot |U|$.
We now demonstrate sufficiency through cases.

\textbf{Case 1.} When $\Hall(G) \in \mathbb{N}$, and $G$ is finite, sufficiency follows from the classical statement of Hall's theorem \cite{Hall1935original} combined with a standard splitting argument. (I.e., creating $\Hall(G)$ many copies of each right-hand side node in the bipartite view of $G$.) 

\textbf{Case 2.} When $\Hall(G) \in \mathbb{N}$ and $G$ is infinite, sufficiency follows from the generalization of Hall's theorem to infinite collections of finite sets, as described in \cite[Theorem 1]{hall1948distinct}.\footnote{We remark briefly that \cite[Theorem 1]{hall1948distinct} appears to be phrased for countable collections of finite sets, but that its proof nevertheless holds for arbitrary collections of finite sets obeying Hall's condition.} 
In particular, each set in the set system corresponds to the collection of edges incident to a given node in $G$, and is thus finite as a consequence of each node's degree being finite.

\textbf{Case 3.} 
When $\Hall(G) = \frac pq \in \mathbb{Q}$, let $G_\bp$ be the bipartite edge-incidence graph of $G$, i.e., $G = (L, R, E')$ with $L = E$, $R = V$, and $(e, v) \in E'$ when $e$ is incident to $v$ in $G$. (See Definition~\ref{defn:bipartite_oig}.) 
Then create a graph $G'_{\bp}$ resembling $G_{\bp}$ but with $q$ copies of each left-hand side vertex, each of which has the same incidence relations as in $G_{\bp}$. By design of $q$, the resulting graph has an assignment such that each right-hand side node receives indegree at least $q \cdot \Hall(G) = p \in \N$. Then interpret each left-hand side node in $G'_{\bp}$ as being in charge of a $\frac 1q$-fraction of an edge in $G_{\bp}$. This gives a fractional/randomized assignment in $G_{\bp}$ for which each right-hand side receives at least $\Hall(G)$ quantity of edges. Interpreting this assignment as an orientation in the original graph $G$, perhaps randomized, yields the desired result. 

\textbf{Case 4.} When $\Hall(G) \not\in \mathbb{Q}$, for any $\epsilon > 0$ there exists $\tfrac{p}{q} \in \mathbb{Q}$ such that $0 < \Hall(G) - \tfrac{p}{q} < \epsilon$. Take $\epsilon \to 0$: for any given $\tfrac{p}{q}$, apply the argument from Case 3. We then obtain a sequence of $\alpha$-orientations for which $\alpha \to \Hall(G)$.
\end{proof}

\begin{definition}
The \defn{Hall complexity} associated to a hypothesis class $\CH$ is the function $\pi_{\CH} : \N \to \N$ defined 
\[ \pi_\CH(n) = \max_{S \in \CX^n} n - \Hall\big(G(\CH|_S)\big). \] 
\end{definition}

\noindent Recall now the transductive error rate of a learner $A$, as defined in Definition~\ref{Definition:trans-learning-error}, i.e., 
\[ \epsilon_{A, \CH}(n) = \max_{S \in \CX^n, h \in \CH} L_{S, h}^{\Trans}(A). \] 
We analagously define the \defn{transductive error rate} of a class $\CH$ as the pointwise minimal error rate attained by any of its learners, i.e., 
\[ \epsilon_{\CH}(n) = \min_{A} \epsilon_{A, \CH}(n). \]

\begin{proposition}\label{Proposition:Hall_seq_main_result}
Fix any domain $\CX$, label set $\CY$, and hypothesis class $\CH \subseteq \CY^\CX$. We have that $\epsilon_\CH(n) = \frac{\pi_\CH(n)}{n}$ for all $n \in \N$. 
\end{proposition} 
\begin{proof}
First note that $\pi_\CH(n)$ is the smallest $\alpha$ for which the graphs $G(\CH|_S)$ are all $\alpha$-coorientable, by Proposition \ref{Proposition:Hall_dimn_orientations} and the fact that each $G(\CH|_S)$ is $\alpha$-coorientable if and only if it is $(n - \alpha)$-orientable. Now, by Lemma~\ref{Lemma:trans_error_coorientations}, $\CH$ has a learner attaining transductive error $\leq \epsilon$ on samples of size $n$ if and only if $G(\CH|_S)$ is $(\epsilon \cdot n)$-coorientable for all $S \in \CX^n$. The claim follows. 
\end{proof}

\begin{remark}
\label{rmk:trans_error_charac_history}
The astute reader may notice the similarity between the Hall complexity, the maximum subgraph density from \citet{haussler1994predicting}, and the maximum average degree from \citet{DS14}. In binary classification, the Hall complexity and the maximum subgraph density are equal, and hence both exactly characterize the transductive error, whereas the maximum average degree is a factor of 2 larger. When there are $k$ labels, the maximum subgraph density serves as a loose lowerbound on the transductive error, and can be up to a factor of $k-1$ smaller. The maximum average degree, on the other hand, serves as an upperbound of the transductive error, and can be up to a factor of 2 larger. The Hall complexity, in exactly characterizing the transductive error, is sandwiched between the maximum subgraph density and the maximum average degree.
We also point out that using Hall's theorem seems to be implicit in the proof of Lemma~57 in \cite{rubinstein2009shifting}, although they are still focused on maximum density, and do not indicate that Hall's theorem permits an exact characterization of transductive error.
\end{remark}

\section{Structural Risk Minimization: Supplement}
\label{Appendix:srm}

\subsection{Regularizer Using $S$ Can Induce Any Learner} \label{Appendix:regularizer_induces_any_learner}

If regularizers are permitted to access the full data of $S$, it is easy to see that the picture degenerates completely: any learner can be witnessed as SRM with respect to a regularizer of this form. In particular, an arbitrary learner $A$ can be recovered as SRM with respect to the following regularizer: 
\begin{align*} 
\psi: \CH \times (\CX \times \CY)^{< \omega} \times \CX &\longrightarrow \R_{\geq 0} \\ 
(h, S, x) &\longmapsto \begin{cases} 0 & A(S)(x) = h(x), \\ 
\infty & A(S)(x) \neq h(x). \end{cases}
\end{align*}
In particular, given a sample $S$ and test point $x$, the regularizer can simply deduce $A(S)$ and output extremely large values for hypotheses disagreeing with $A(S)$ on $x$. Note that this argument relies crucially upon the regularizer's access to the test point in $\CX$ as an input; any regularizer which is uniform with respect to $\CX$ (even if granted full information of $S$) begets proper learners, which are insufficient by Proposition \ref{Proposition:proper_fails_multiclass}.

Note also that the guarantee $A(S)(x) \in \CH(x)$ holds for all $x \in \CX$ when $A$ is any sensible learner, as we are learning in the realizable case with respect to the 0-1 loss. (In particular, any learner $A$ violating this condition of ``local properness" can be improved by obligating it to be locally proper, in any way.) Further, as our losses are bounded by 1, a regularizer's output of $\infty \notin \R_{\geq 0}$ in this example is equivalent to simply outputting $c$ for any $c > 1$.

\subsection{Support for Conjecture~\ref{Conjecture:size-based-regularizer-fails}}
\label{Appendix:support-for-conjecture}

We now present a candidate hypothesis class $\CH$ which may justify our Conjecture~\ref{Conjecture:size-based-regularizer-fails}. That is, we will define a class $\CH$ which is PAC learnable, and for which we suspect that $\CH$ \emph{cannot} be learned by a local size-based regularizer. Let $\CX$ be an infinite set, say $\CX = \N$, and $\CY = \{*\} \cup 2^{\CX}$, where we use $2^\CX$ to denote the power set of $\CX$. 

Before defining $\CH$, let $\CJ \subseteq (2^\CX)^3$ be the collection of all triples of subsets $(R, S, T)$ such that: 
\begin{itemize}
    \item[1.] $|R| = |S| = |T| =: k \in 2 \N$
    \item[2.] $| R \cap S| = |R \cap T| = |S \cap T| = k / 2$. In particular, $R \cap S \cap T = \emptyset$. 
\end{itemize}
For each $(R, S, T) \in \CJ$, we will define 8 hypotheses in $\CH$. Namely, all those hypotheses $h \in \CY^\CX$ satisfying: 
\begin{itemize}
    \item[1.] $h(x) = *$ for all $x \notin R \cup S \cup T$. 
    \item[2.] $h$ is constant on each of $R \cap S$, $R \cap T$, and $S \cap T$. 
    \item[3.] For $x \in R \cap S$, $h(x) \in \{R, S\}$; for $x \in R \cap T$, $h(x) \in \{ R, T\}$; and for $x \in S \cap T$, $h(x) \in \{S, T\}$. 
\end{itemize}
Informally, each such $h$ is simply the constant function $\_ \mapsto *$ outside of $R \cup S \cup T$, and on the regions $R \cap S$, $R \cap T$, and $S \cap T$ has the choice of acting as a constant function taking a value in $\{R, S\}$, $\{R, T\}$, or $\{S, T\}$ respectively. Geometrically, one can think of such a function as choosing how to layer the regions $S$, $T$, and $R$ ontop of each other. (I.e., sheets of paper over each of $S$, $T$, and $R$ bearing the names of their corresponding sets; a function $h$ is equivalent to a choice of layering the sheets of paper with respect to each other. The output of $h$ at input $x$ is the label at $x$ seen from above, i.e., of the topmost sheet of paper.) 

We define the class $\CH$ to be the union of all such functions over all triples $(R, S, T) \in \CJ$. One can see that that $\CH$ is PAC learnable by the learner which defaults to outputting $*$ at $x \in \CX$ unless the label $S \ni x$ has been observed in the sample, in which case it outputs $S$.\footnote{If \emph{two} such labels $S \ni x$ and $R \ni x$ have been seen in the training sample, and this information reveals the true label of $x$ (i.e., one label was seen on $S \cap T$), then simply predict this label. If two such labels were seen but this does not reveal the true label of $x$, arbitrarily choose either of $S$ or $T$).} Informally, consider any function $h$ arising from an $(R, S, T) \in \CJ$ and a realizable distribution $D$ with marginal $D_\CX$ over $\CX$. Then the previous learner incurs true error 0 once unlabeled datapoints have been seen in each of $R \cap S$, $S \cap T$, and $T \cap R$. Any such region either has negligible mass under $D_\CX$ or will quickly be observed in a training set. 

Let us now argue why we suspect $\CH$ not to be learnable by a local size-based regularizer. First note that ERM learners fail miserably to learn $\CH$: fix a large set $S \subseteq \CX$ and let $D$ be the uniform distribution over points of the form $\{(x, S) : x \in S\}$. Then $D$ is a realizable distribution, and consider the output of an ERM learner on a training set $\Strain \sim D$ with $|\Strain| < |S| / 2$. With probability $> 1/2$, a test point $(\xtest, S) \sim D$ will be such that $\xtest$ was not seen in $\Strain$. In this case, there exists a hypothesis $h \in \CH$ with empirical error 0 such that $h(\xtest) \neq S$. Namely, an $h$ arising from a triple of sets $(R, S, T)$ such that $\Strain \cap \big( R \cap S \big) = \emptyset$ and $\xtest \in R \cap S$. An ERM learner is free to predict label $R$ at such an $\xtest$, there incurring constant error at test time. As the original set $S$ may be chosen to be arbitrarily large, the problem affects ERM learners trained on arbitrarily large training sets $\Strain$. 

Informally, any learner $\CA$ equipped with only the information of $\xtest$ and the empirical errors of all $h \in \CH$ would seem to suffer from such a problem on uniform distributions over $\{(x, S): x \in S\}$. That is, the great amount of symmetry inherent in $\CH$ prevents $\CA$ from recognizing that $S$ is the most ``natural" prediction for $\xtest$, in contrast to any of the sets $R$ which contain $\xtest$ yet avoid $\Strain$. In short, it seems that a learner must peek into the training set $\Strain$ in order to learn the geometry of the underlying distribution and have a chance of learning. Local size-based regularizers, however, are not permitted to peek into the training set.

\section{Omitted proofs}
\label{appx:omitted_proofs}

\subsection{Proof of Proposition~\ref{Proposition:local_regularizer_fails_multiclass}}
\label{appendix:local_regularizer_fails}
\begin{proof}
We use the \emph{first Cantor class} of \citep{daniely-et-al2011,DS14}. In particular, let $\{\CX_d\}_{d \in 2\N}$ be a disjoint collection of sets with $|\CX_d| = d$. Furthermore, let $\CY_d = 2^{\CX_d} \cup \{ * \}$ for each $d \in 2\N$. For each $A \subseteq \CX_d$, define $h_A : \CX_d \to \CY_d$ by 
\[ h_A(x) = \begin{cases} A & x \in A, \\ * & x \notin A. \end{cases} \]
Now let $\CX_\infty = \cup_{d \in 2\N} \; \CX_d$ and $Y_\infty = \big(\cup_{d \in 2\N} 2^{\CX_d} \big) \cup \{*\}$. We can extend each $h_A : \CX_d \to \CY_d$ to a function $\CX_\infty \to \CY_\infty$ by simply definining it to return $*$ outside of $\CX_d$. Lastly, set 
\[ \CH_\infty = \Big\{ h_A : A \subseteq \CX_d \text{ for some $d$}, |A| = \frac d2 \Big\}. \] 
Then $\CH_\infty$ is PAC learnable as a consequence of \cite[Lemma 20]{DS14}, i.e., by the learner that returns the constant function $*$ if it attains zero empirical risk and the unique $h_A$ attaining zero empirical risk otherwise. Note that this constant function is not in $\CH_\infty$. 

We now show that $\CH_\infty$ cannot be learned by any local SRM learner. Fix a local regularizer $\psi: \CH \times \CX \to \R_{\geq 0}$. Now define $\psi_\CH : \CX \times \CY \to \R_{\geq 0}$ as follows,
\[ \psi_\CH(x, y) = \inf_{\substack{h \in \CH, \\ h(x) = y}} \psi(h, x). \] 
Informally, $\psi_\CH$ captures the local preferences over labels, rather than entire hypotheses, induced by $\psi$. We will say that $\psi$ weakly prefers $y$ to $y'$ at $x$ if $\psi_\CH(x, y) \leq \psi_\CH(x, y')$. 

Suppose now that the following property $P$ holds of $\psi$: for each $x \in \CX_\infty$, $\psi$ weakly prefers every set $A$ containing $x$ to $*$. We show that there is a learner $\CA$ induced by $\psi$ that is not a PAC learner for $\CH_\infty$. Pick some $A \subseteq \CX_d$ of size $\frac d2$. Let $D_{A}$ be the uniform measure over the finite set $\big\{(x, *) : x \in \CX_d \setminus A \big\}$. Note that $D_{A}$ is an $\CH_\infty$-realizable distribution, as $h_{A} \in \CH_\infty$ attains a true error of zero with respect to it. Now let $S \sim D_A^m$ be a sample of size $m < d/4$. As $|\CX_d \setminus A| = \frac d2$, $S$ does not contain the full support of $D_A$. In particular, for any $x \in \CX_d \setminus S$, there exists an $A_x \subseteq \CX_d$ of size $\frac d2$ such that $x \in A_x$ and $A_x \cap S = \emptyset$. 

For such $x$, $L_S(A_x) = 0$, as $A_x$ avoids $S$. Furthermore, by definition of property $P$, $\psi_\CH(x, A_x) \leq \psi_\CH(x, *)$, meaning $\psi(h_{A_x}, x) \leq \psi(h_B, x)$ for every $B$ such that $x \notin B$. Thus $\CA$ can be taken such that $\CA(S)(x) \neq *$. As this holds for all $x \in \CX_d \setminus S$, $\CA$ misclassifies all points in $\CX_d \setminus A$ that are not in $S$. Consequently, $A$ incurs an expected true error of at least $\frac 12$. And $d$ can be taken to be arbitrarily large by ranging across $\{\CX_d\}_{d \in 2\N} \subseteq \CX_\infty$, meaning $\CA$ is not a PAC learner. 

Suppose now that $\psi$ does not satisfy property $P$. Then there exists an $x \in \CX_d$ and $A \ni x$ such that $\psi_{\CH}(x, A) > \psi_{\CH}(x, *)$. In particular, as $h_A$ is the only function in $\CH_\infty$ with $A$ in its image, we have $\psi(h_A, x) > \psi(h_B, x)$ for any $B$ such that $x \notin B$. Set $c_2 = \psi_{\CH}(x, A)$ and $c_1 = \psi_{\CH}(x, *)$. Now let $D$ be the distribution placing $\min(\frac{c_2 - c_1}{2}, 1)$ mass on the point $(x, A)$ and its remaining mass uniformly across datapoints in $\CX_d \setminus A$ with label $*$. For increasingly large samples $S$ drawn from $D$, the proportion $p_S$ of sample points taking the form $(x, A)$ will satisfy $p < c_2 - c_1$ with high probability. Consequently, at point $x$, $h_A$ will attain greater structural risk than some $h_B$ with $x \notin B$. That is, $\CA(S)(x) \neq A$, and thus $\CA$ incurs constant true error over arbitrarily large samples. This completes the proof. 
\end{proof}

\subsection{Proof of Theorem~\ref{Theorem:unsupervised-local-srm-factor2}}
\label{appx:proof-srm-factor-2}
Before commencing with the proof, it will be useful to establish a basic fact: local regularizers of any kind should serve only to ``tie-break" between hypotheses attaining zero empirical risk. 

\begin{lemma}\label{Lemma:feature-SRM-1/|S|-tiebreaking}
Let $\psi: \CH \times \CX^{<\omega} \times \CX \to \R_{\geq 0}$ be a local unsupervised regularizer. Then, without loss of generality, $\psi$ can be assumed to obey the following property for all $S \in \CX^n$ and $x \in \CX$: 
\[ \psi(h, S, x) < \frac{1}{|S|} \quad \forall h \in \CH.  \] 
\end{lemma}
\begin{proof}
As we are learning in the realizable case with respect to the 0-1 loss, any learner $A$ can be assumed to satisfy $A(S)(x) \in \big\{ h(x) : h \in \argmin_{\CH} L_S(h) \big \}$. In particular, for the underlying distribution $D$, there exists an $h \in \CH$ with $L_D(h) = 0$. Then on any sample $S$, $h \in \argmin_\CH L_S(h)$ with probability 1. If $A$ violates the previous property, then with probability 1, $A(S)(x) \neq h(x)$ and thus $\ell_{0-1}\big(A(S)(x), h(x)\big) = 1$, meaning $A$ incurs the maximal loss possible (for any underlying distribution $D$). 

Consequently, any regularizer $\psi$ should be such that 
\begin{align*}
  \Big\{ h(x) : h \in \argmin_{\CH} L_S(h) + \psi(h, S_\CX, x) \Big\} &\subseteq \Big\{ h(x) : h \in \argmin_{\CH} L_S(h) \Big \} \\
  &= \Big\{ h(x) : L_S(h) = 0 \Big \}. 
\end{align*}
In particular, the action of $\psi$ should only help compare hypotheses attaining equal empirical risk. As $L_S(h) \in \{\frac{1}{|S|}, \ldots, \frac{|S|}{|S|}\}$ for any $h \in \CH$, it suffices to show that an arbitrary $\psi$ can be compressed to the interval $[0, \frac{1}{|S|})$ by a strictly increasing function. And indeed this can be achieved, by such a function as $x \mapsto  \frac{2}{\pi |S|} \tan^{-1}(x)$. 
\end{proof}

Perhaps the most important ingredient to our central result is the relationship between unsupervised regularizers and acyclic orientations of OIGs. In particular, to demonstrate that an OIG can be oriented favorably by an unsupervised regularizer (i.e., by its induced learners), it suffices to demonstrate that it can be oriented favorably in an acyclic manner. 

\begin{proposition}\label{Proposition:acyclic-same-as-unsupervised-SRM}
Let $\CY$ be a finite or countable label set, $\CH \subseteq \CY^\CX$ a hypothesis class, and $A$ a learner for $\CH \subseteq \CY^\CX$, corresponding to a collection of orientations for the one-inclusion graphs $\{G(\CH|_S)\}_{S \in \CX^{<\omega}}$. If $A$ begets acyclic orientations on all the one-inclusion graphs of $\CH$, then it is a local unsupervised SRM learner for some $\psi$. 
\end{proposition}
\begin{proof}
Suppose that $\CY$ is finite and that $A$ gives rise to acylic orientations of all one-inclusion graphs for $\CH$. We claim that a local unsupervised regularizer for $\CH$ can be thought of as an arbitrary function $\phi: V_\infty \to \R_{\geq 0}$, for $V_\infty$ the union of vertices across all the one-inclusion graphs of $\CH$. This can easily be seen by observing that a local unsupervised regularizer $\psi$ can choose to judge each $h \in \CH$ based only upon the information of its restriction to $S_\CX \cup \{x_{\text{test}}\}$. Now fix one such graph $G(\CH|_S) = (V, E)$ for $S \in \CX^n$, directed by the action of $A$. As it is acyclic, it can be topologically ordered, so that each vertex $v \in V$ lies in layer $\ell_v \in \N$. 

It thus suffices to exhibit a function $\phi : V \to \R_{\geq 0}$ such that $A$ orients each $e \in E$ toward an incident vertex with maximal output under $\phi$. This is achieved by, for instance, 
\[ \phi: v \mapsto \frac{1}{2|S|} \cdot \left ( 1-\frac{1}{\ell_v} \right). \] 
Notably, $\phi$ satisfies each of the two properties we require of it: that it be strictly increasing with nodes' layers in the topological ordering of $G(\CH|_S)$, and that it be bounded above by $\frac{1}{2|S|}$, in accordance with Lemma \ref{Lemma:feature-SRM-1/|S|-tiebreaking}. 

The case for countable $\CY$ is slightly more involved. 
Let $\CY$ be countably infinite and $A$ a learner for $\CH \subseteq \CY^\CX$  that induces acyclic orientations on all the one-inclusion graphs of $\CH$, $\{G(\CH|_S)\}_{S \in \CX^{< \omega}}$. Fix one such $S \in \CX^n$ and $G(\CH|_S) = (V, E)$ the accompanying one-inclusion graph. Note that $V$ is at most countable, as it has cardinality at most $|\CY^n| = |\CY|$. The acyclic orientation on $G$ endows its vertex set $V$ with a partial order $P$, defined by $u \leq v$ if $u = v$ or if there exists a path $p : u \to v$. 

Appealing to Zorn's lemma, $P$ can be completed into a total order on $V$. Furthermore, $V$ then embeds into $\R$ as a totally ordered set owing to its countability. That is, one can embed $V = \{v_i\}_{i \in \N}$ into $\R$ inductively, beginning with $v_0 \mapsto 0$. Once $(v_i)_{i < k}$ have been embedded as $(r_i)_{i < k}$, $v_k$ can be mapped to $\max_{i < k} r_i + 1$ if $v_k > \max_{i < k} v_i$, to $\min_{i < k} r_i - 1$ if $v_k < \min_{i < k} v_i$, and otherwise to $\frac{r_i + r_j}{2}$ for $v_i$ and $v_j$ the least upper bound and greatest lower bounds of $v_k$ in $(v_i)_{i < k}$. 

Lastly, by post-composing such an embedding with an isomorphism of ordered sets $\R \to (0, \frac{1}{|S}|)$, such as $x \mapsto \frac{2}{\pi |S|} \tan^{-1}(x)$, we have that $V$ embeds into $(0, \frac{1}{{S}})$ as a totally ordered set. Call this embedding $\phi$. As in the proof for finite $\CY$, $\phi$ gives rise to a local unsupervised regularizer $\psi$ that recovers precisely the action of $A$. That is, $\psi(h, S, x)$ chooses to judge $h$ based upon its restriction to $S \cup \{x\}$, at which point it acts as the embedding $\phi$. 

Notably, $\psi$ satisfies the two crucial properties required of it: that it be bounded above by $\frac{1}{|S|}$, in accordance with Lemma~\ref{Lemma:feature-SRM-1/|S|-tiebreaking}, and that it be compatible with the acyclic orientation of $G(\CH|_S)$ induced by $A$. 
\end{proof}

We are now ready to prove Theorem~\ref{Theorem:unsupervised-local-srm-factor2}. 

\begin{proof}
By Proposition \ref{Proposition:acyclic-same-as-unsupervised-SRM}, it suffices to demonstrate the following claim: for any $S \in \CX^n$, there exists an acylic orientation of $G(\CH|_S)$ that is optimal within a factor of 2, i.e., that minimizes nodes' maximal outdegrees to within a factor of 2. In particular, note that the unsupervised regularizer $\psi$ described in Proposition \ref{Proposition:acyclic-same-as-unsupervised-SRM} induces a unique learner. We will use the \emph{density} of a finite undirected graph to refer to the average degree of its nodes. Now suppose that $G(\CH|_S)$ is $\alpha$-coorientable, as described in Definition \ref{Definition:alpha-(co)orientable}. Then the maximal density of any finite subgraph of $G(\CH|_S)$ is at most $2 \alpha$, by \cite[Theorem 2]{DS14}. When $G(\CH|_S) = (V, E)$ is finite, this allows us to design an acyclic $(2 \alpha )$-coorientation by employing the following \emph{$k$-core} algorithm implicit in \cite[Lemma 3]{DS14}: for $i \in [|V|]$, compute the vertex $v \in G(\CH|_S)$ of lowest degree, remove it from $G(\CH|_S)$, and place it in the $i$th layer of a topological ordering of $G(\CH_S)$. The outdegree of any vertex in this topological ordering is precisely its degree in the undirected graph $G(\CH|_S)$ before it was removed, which is at most $2 \alpha$ owing to the maximal subgraph density of $G(\CH|_S)$. 

When $G(\CH|_S)$ is infinite, the claim follows from the compactness theorem of propositional calculus (see \cite{DS14, brukhim2022characterization}), which, for completeness, we detail as follows.

Let $G(\CH|_S) = (V, E)$ be infinite and $\beta$ the maximal density of any of its finite subgraphs (i.e., the average degree of nodes). By \cite[Theorem 2]{DS14}, it suffices to show that $G(\CH|_S)$ can be $\beta$-cooriented. Let $|S| = n$, and for any $(v, e) \in V \times E$ let $E_v$ denote the set of all edges incident to $v$ and $V_e$ the set of all vertices incident to $e$. For vertices $u, v \in V$, let $\CP_{u, v}$ denote the set of all paths from $u$ to $v$, i.e., of finite sequences $p = \big((e_1, v_1), \ldots, (e_\ell, v_\ell) \big)$ such that $e_1$ is incident to $u$, $e_i$ is incident to each of $v_{i}$ and $v_{i-1}$ (when they exist), and $v_\ell = v$. 

Now consider the set of propositional variables $\{P_{e, v} : e \in E, v \in V_e\}$. Intuitively, $P_{e, v}$ will be true if the edge $e$ is oriented to $v$ and false otherwise. We define a set of sentences $\Sigma$ as follows. 
\begin{enumerate}
    \item $\lnot (P_{e, v} \land P_{e, v'})$ for all $e \in E$ and pairs of nodes $v, v'$ both incident to $e$. 
    \item $\displaystyle \bigvee_{\substack{E'_v \subseteq E_v, \\ |E'_v| \geq n - \beta}} \bigwedge_{e' \in E'_v} P_{e', v}$ for all $v \in V$. 
    \item $ \displaystyle \left( 
    \lnot \bigwedge_{(e, s) \in p} P_{e, s} \right) \lor \left( \lnot \bigwedge_{(e', s') \in p'} P_{e', s'} \right)$ for all $u, v \in V$ and $(p, p') \in \CP_{u, v} \times \CP_{v, u}$.  
\end{enumerate}
Sentences from (1.)\ correspond to the requirement that each edge be oriented to at most one incident vertex, those from (2.)\ demand that each node have indegree at least $n - \beta$ (equivalently, outdegree at most $\beta$), and those from (3.)\ demand that $G(\CH|_S)$ never contain both a path $u \to v$ and $v \to u$ (i.e., a cycle). Note that sentences from (2.)\ are finite because each vertex $v$ has finite degree exactly $n$ in the undirected graph $G(\CH|_S)$, and sentences from (3.)\ are finite because paths in $\CP_{u, v}$ are finite. 

Let us now demonstrate that $\Sigma$ is finitely satisfiable. Suppose we impose a finite collection of sentences $\Sigma' \subseteq \Sigma$, involving only the variables $P_{e, v}$ for the finite set $V' \subseteq V$. Let $G[V'] \subseteq G(\CH|_S)$ be the subgraph of $G(\CH|_S)$ with vertex set $V'$ and edge set $E' \subseteq E$ consisting of those edges with at least two incident nodes in $V'$. As $G[V']$ is a finite subgraph of $G(\CH|_S)$, its density is at most $\beta$. By our previous work from the proof of Theorem \ref{Theorem:unsupervised-local-srm-factor2} for finite $\CY$, $G[V']$ can $\beta$-cooriented acyclically. Such an acyclic $\beta$-coorientation amounts precisely to a choice of $P_{e, v}$ for each $(e, v) \in E' \times V'$ satisfying sentences (1.),\ (2.),\ and (3.)\ restricted to $G[V']$. Setting all the remaining variables to false continues to satisfy $\Sigma'$, thus demonstrating that $\Sigma$ is finitely satisfiable. 

Then, by compactness, $\Sigma$ is satisfiable, meaning there exists a partial orientation $\sigma$ of $G(\CH|_S)$ such that each node has outdegree at most $\beta$ and there are no directed cycles. In particular, there may be an $e \in E$ such that $P_{e, v}$ is false for all $v \in V_e$. Such a partial orientation on $G(\CH|_S)$ endows its vertex set $V$ with a partial order $P$, defined by $u \leq v$ if $u = v$ or if there exists a directed path $p : u \to v$. Appealing to Zorn's lemma, $P$ can be completed into a total order on $V$. Embedding $V$ into $(0, \frac{1}{|S|})$ as a totally ordered set --- as in the proof of Proposition \ref{Proposition:acyclic-same-as-unsupervised-SRM} --- defines a regularizer $\psi$. Notably, $\psi$ recovers the action of $\sigma$ on the respective edges, and any manner of completing $\sigma$ into a total orientation can only increase nodes' indegrees (i.e., reduce their outdegrees), completing the argument. 
\end{proof}

\begin{remark}
The unsupervised regularizer employed in Propositions \ref{Proposition:acyclic-same-as-unsupervised-SRM} and \ref{Theorem:unsupervised-local-srm-factor2} does not distinguish between $x_{\test}$ and the elements of $S_\CX$. In particular, it is symmetric with respect to $x_{\test}$ and any element of $S_\CX$. This has two central consequences: 
\begin{enumerate}
    \item Propositions \ref{Proposition:acyclic-same-as-unsupervised-SRM} and \ref{Theorem:unsupervised-local-srm-factor2} hold for a regularizer that factors through $S_\CX \cup \{x_\test \}$, i.e., that may as well be defined to receive $S_\CX \cup \{x_\test\}$ as input. This demonstrates sufficiency of a regularizer receiving even less information than the local unsupervised regularizers of Definition~\ref{Definition:local-unsupervised-regularizer}. 

    \item Semantically, this implies the existence of near-optimal transductive learners based on regularizers that decide their values in the transductive setting merely after observing the collection of unlabeled datapoints (including test point). Furthermore, it demonstrates that an OIG $G = (V, E)$ can always be oriented near-optimally by assigning a value to each node $v \in V$, rather than by assigning an incident node to each $e \in E$. This achieves another central aim of this section: to describe optimal orientations of OIGs more parsimoniously, using global structure rather than local structure. 
\end{enumerate}
\end{remark}

\subsection{Proof of Theorem~\ref{thm:realizable_randomized_learner}}
\label{appx:proof_max_entropy}

Before proving Theorem~\ref{thm:realizable_randomized_learner}, we introduce the notion of an \emph{assignment} for a bipartite graph $G$, which will replace our discussion of orientations in $G(\CH|_S)$, as the two are equivalent.
\begin{definition}
Let $G = (\CA, \CB, E)$ be a bipartite graph. An \edefn{assignment} in $G$ is a function $\sigma: \CA \to \CB$ such that $(a, \sigma(a)) \in E$ for all $a \in \CA$. 
\end{definition}

\noindent The following is immediate from the definitions of OIGs and their bipartite counterparts. 

\begin{lemma}\label{lemma:orientations_and_assignments}
Fix a hypothesis class $\CH \subseteq \CY^\CX$ and sequence $S \in \CX^n$. Let $G(\CH|_S)$ be the OIG of $\CH$ with respect to $S$ and $G_{\bp} = (\CA, \CB, E)$ its bipartite counterpart, as defined in Definition~\ref{defn:bipartite_oig}. Then the following are equivalent for any $\alpha \in \N$. 
\begin{enumerate}
\setlength\itemsep{0 em}
\item There exists an $(n - \alpha)$-orientation of $G(\CH|_S)$; and
\item There exists an assignment $\sigma$ of $G_{\bp}$ such that each $v \in \CB$ receives at least $n-\alpha$ degree, i.e., $|\sigma^{-1}(v)| \geq n - \alpha$.
\end{enumerate}
\end{lemma}

With the language of assignments, we can begin discussion of our maximum entropy convex program.
\begin{proposition}
\label{prop:optimality_of_max_entropy}
Let $\CH$ be a hypothesis class, and
recall the optimal transductive error rate $\epsilon_{\CH}(n) = \frac{\pi_\CH(n)}{n}$.
For any $S \in \CX^n$, let $G(\CH|_S)$ be the one-inclusion graph of $\CH$ on $S$ and $\CD$ the collection of all orientations of $G(\CH|_S)$ (equivalently, the collection of assignments in $G_{\bp} = (\CA, \CB, E)$, the bipartite analogue of $G(\CH|_S)$). 
Then there is a unique distribution over assignments $P^* \in \Delta_\CD$ such that each $v \in \CB$ receives at least $n \cdot (1 - \epsilon_{\CH}(n)) = n - \pi_\CH(n)$ in-degree in expectation, and the following conditions hold simultaneously:
\begin{enumerate}
    \item $P^*$ achieves maximum entropy among all $P \in \Delta_\CD$ subject to the degree requirements.
    \item $P^*$ induces a randomized transductive learner $\CA$ which, in expectation over the randomness of the learner, achieves optimal error rate $\epsilon_{\CH}(n)$.
\end{enumerate}
\end{proposition}
\begin{proof}
(2.)\ follows immediately from the connection between in-degree and transductive error, as established in Lemma~\ref{Lemma:trans_error_coorientations}. 

(1.) is proven using the maximum entropy convex program of \cite{singh2014entropy}, with a slight modification owing to the fact that we are in a more general multiset setting.
Let $\CD$ be the set of all assignments in $G_{\bp}$. Recall that such assignments can be thought of as learners, as discussed in Section~\ref{subsec:OIG}.
We will think of each $d \in \CD$ as a $|\CA| \times |\CB|$ matrix with 0-1 entries, where all row sums are exactly 1 (that is, every partially labeled dataset in $\CA$ is oriented toward one fully labeled dataset in $\CB$).
We will index entries of this matrix as $d(u, v)$ for $u \in \CA, v \in \CB$.

Let $n$ be fixed.
Our goal is to find a maximum entropy distribution over assignments $P^* \in \Delta_\CD$ such that each $v \in \CB$ is
allocated at least $c = (1 - \epsilon_{\CH}(n))n$ in-degree in expectation.
By Lemma~\ref{Lemma:trans_error_coorientations}, this would suffice to define a (randomized) learner induced by $P^*$ which achieves error at most $\epsilon_{\CH}(n)$.

Therefore, we want the solution to the following maximum entropy optimization problem, which we term $\maxent$.
\begin{align*}
    \maxent = \underset{p_d:\ d \in \CD}{\text{max}} \quad & \sum_{d \in \CD} p_d \ln \frac{1}{p_d} \\
    \text{s.t.} \quad & \sum_{d \in \CD} p_d \sum_{u \in \CA} d(u, v) \geq c
    \quad &\forall v \in \CB \\
    & \sum_{d \in \CD} p_d = 1\\
    & p_d \geq 0 \quad &\forall d \in \CD
\end{align*}
    Notice that $\maxent$ is a convex program with a concave objective and linear constraints.
The number of variables is equal to the number of assignments in the graph $G_{\bp}$.

Let us confirm that the program is feasible and, in fact, enjoys a unique optimal solution. Feasibility arises as a consequence of the fact that some randomized learner attains error rate $\epsilon_{\CH}(n)$, corresponding to a randomized orientation of the OIG that satisfies the degree bounds required by our program. Uniqueness of the solution follows from strict concavity of the objective.
This completes the proof of (1.).
\end{proof}

Theorem~\ref{thm:realizable_randomized_learner} demonstrates that the maximum entropy learner can furthermore be interpreted  as \emph{Bayesian} in the following sense: First, it learns a prior over hypotheses from the unlabeled data. (In particular, over the projection of $\CH$ to the unlabeled data.) Second, given the labels of all but the test point, it performs a Bayes update to form a posterior over the hypotheses consistent with these labels. Third, it samples a hypothesis from this posterior, which it then uses to predict a label for the test point.

The proof of Theorem~\ref{thm:realizable_randomized_learner} will build upon  the dual characterization of the maximum entropy convex program $\maxent$ defined in Proposition~\ref{prop:optimality_of_max_entropy}.
To ensure that this program (and its dual) are well-defined, we need to check three things: feasibility, uniqueness of the optimal solution, and strong duality.
The proof of Proposition~\ref{prop:optimality_of_max_entropy} already addresses feasibility and uniqueness.
Strong duality follows from the following observation: Notice that the inequality constraints of $\maxent$ are affine (linear) in the variables.
Therefore, we can directly apply the weak version of Slater's condition to get that strong duality holds.
Recall that weak Slater's condition requires strict feasibility only in non-affine inequality constraints (of which we have none).
Then, feasibility is enough to prove strong duality.

The final key ingredient to the proof of Theorem~\ref{thm:realizable_randomized_learner} is the following lemma which relates the (optimal) primal and dual variables of $\maxent$.
\begin{lemma}
\label{lemma:dual-variables-decomposition}
    Let $\lambda_v \in \R$ for all $v \in \CB$ and $z \in \R$ be the optimal dual variables.
    Then, for each $d \in \CD$, the associated optimal primal variable $p_d$ can be written as $p^*_d = e^{-1 - z} \prod_{(u,v) \in d} \exp(-\lambda_v)$.
\end{lemma}
\begin{proof}
Since strong duality of $\maxent$ holds, we will derive the dual problem similar to \cite{singh2014entropy} (see their Appendix A.1).
First, we find the Lagrangian.
\begin{align}
    L(p, \lambda, z)
    &= \sum_{d \in \CD} p_d \ln(\frac{1}{p_d}) + \sum_{v \in \CB} \lambda_v (c - \sum_{d \in \CD} p_d \sum_{u \in \CA} d(u,v)) + z(1 - \sum_{d \in \CD} p_d) \\
    &= \sum_{d \in \CD} p_d \ln(\frac{1}{p_d}) + c\sum_{v \in \CB}\lambda_v - \sum_{d \in \CD} p_d \sum_{v \in \CB} \lambda_v \sum_{u \in \CA} d(u,v) + z - z\sum_{d \in \CD} p_d
    \label{eq:plug-into}
\end{align}
Now take partials and set to zero.
\begin{align}
    \frac{\partial L}{\partial p_d} = p_d \cdot \frac{1}{\frac{1}{p_d}} \cdot \frac{\partial}{\partial p_d} (\frac{1}{p_d}) - \sum_{v \in \CB} \lambda_v \sum_{u \in \CA} d(u,v) - z + \ln(\frac{1}{p_d}) &= 0\\
    -1 - \sum_{v \in \CB} \lambda_v \sum_{u \in \CA} d(u,v) - z + \ln(\frac{1}{p_d}) &= 0
    \label{eq:mult_pA}
\end{align}
Therefore,
\begin{align}
    \ln(\tfrac{1}{p_d}) &= 1 + \sum_{v \in \CB} \lambda_v \sum_{u \in \CA} d(u,v) + z \nonumber \\
    p_d &= \exp\left(-1 - \sum_{v \in \CB} \lambda_v \sum_{u \in \CA} d(u,v) - z\right) \label{pA_exp}.
\end{align}
Summing over all $d \in \CD$,
\begin{align*}
    \sum_{d \in \CD} p_d = e^{-1 - z} \sum_{d \in \CD} \exp \left( - \sum_{v \in \CB} \lambda_v \sum_{u \in \CA} d(u,v)\right).
\end{align*}
This obtains the characterization of the probability of an assignment in terms of the dual variables, as we required.

For completeness, we finish the derivation of the dual optimization problem.
Multiply each \eqref{eq:mult_pA} by $p_d$ and sum over all $A$ to get that
\begin{align*}
    \sum_{d \in \CD}\left( -p_d - p_d \sum_{v \in \CB} \lambda_v \sum_{u \in \CA} d(u,v) - z p_d + p_d \ln(\tfrac{1}{p_d}) \right) = 0,
\end{align*}
implying that
\begin{align*}
    \sum_{d \in \CD} p_d = \sum_{d \in \CD} \left(- p_d \sum_{v \in \CB} \lambda_v \sum_{u \in \CA} d(u,v) - z p_d + p_d \ln(\tfrac{1}{p_d}) \right).
\end{align*}
Plug these two facts into \eqref{eq:plug-into}:
\begin{align*}
    L(p, \lambda, z) = c\sum_{v \in \CB} \lambda_v + z + \sum_{d \in \CD} p_d
    = c\sum_{v \in \CB} \lambda_v + z + e^{-1 - z} \sum_{d \in \CD} \exp \left( - \sum_{v \in \CB} \lambda_v \sum_{u \in \CA} d(u,v)\right).
\end{align*}
Which is only a function of $\lambda, z$.
Take partial with respect to $z$ and set to zero:
\begin{align*}
    \frac{\partial L}{\partial z} = 1 - e^{-1 - z} \sum_{d \in \CD} \exp \left( - \sum_{v \in \CB} \lambda_v \sum_{u \in \CA} d(u,v)\right) &= 0\\
    z = -1 + \ln\left( \sum_{d \in \CD} \exp \left( - \sum_{v \in \CB} \lambda_v \sum_{u \in \CA} d(u,v)\right)\right).
\end{align*}
Therefore, the dual optimization problem then becomes:
\begin{align*}
    \min_{\forall v \in \CB,\ \lambda_v \in \R} c \sum_{v \in \CB} \lambda_v + \ln \sum_{d \in \CD} \exp \left( - \sum_{v \in \CB} \lambda_v \sum_{u \in \CA} d(u,v)\right).
\end{align*}
Notice that \eqref{pA_exp} gives us the optimal value for each $p_d$:
\begin{align*}
    p_d = e^{-1 - z} \exp\left( -\sum_{v \in \CB} \lambda_v \sum_{u \in \CA} d(u,v) \right)
    = e^{-1 - z} \prod_{v \in \CB} \prod_{u \in \CA} \exp\left(-\lambda_v d(u, v)\right)
    = e^{-1 - z} \prod_{(u,v) \in d} \exp(-\lambda_v),
\end{align*}
where the last equality follows from the fact that $d(u,v) \neq 0$ only for $(u,v)$ which exist in the assignment $d$.
\end{proof}

We now prove Theorem~\ref{thm:realizable_randomized_learner}. 

\begin{proof}
    Let $G_{\bp} = (\CA, \CB, E)$ be the bipartite OIG associated to the unlabeled datapoints $S_\CX^+ = (x_1, \ldots, x_n)$. Let $\CD$ denote the collection of assignments in $G_{\bp}$. 
    For a given assignment $d \in \CD$, we say that $(u, v) \in d$ if the edge $(u,v) \in E$ is selected in $d$, and write $d(u,v) = 1$.
    From the dual derivation of $\maxent$ in Lemma~\ref{lemma:dual-variables-decomposition}, within the optimal randomized maximum entropy learner $P^*$, each $d \in \CD$ has an associated probability given by:
    \begin{align*}
        p^*_d = e^{-1 - z} \prod_{(u,v) \in d} \exp(-\lambda_v)
    \end{align*}
    where $\lambda_v$ are dual variables corresponding to each $v \in \CB$, and the final dual variable $z$ has identical value for each $d \in \CD$ given by:
    \[z = -1 + \ln\left( \sum_{d \in \CD} \exp \left( - \sum_{v \in \CB} \lambda_v \sum_{u \in \CA} d(u,v)\right)\right).\]
    For all $v \in \CB$, let $\gamma_v = \exp(-\lambda_v)$, and define $\rho_v = \gamma_v / \sum_{v \in \CB} \gamma_v$.
    Then we can rewrite $p^*_d$ as follows.
    \begin{align*}
        p^*_d = e^{-1 - z} \prod_{(u,v) \in d} \gamma_v 
        &= e^{-1-z} \left(\sum_{v \in \CB} \gamma_v\right)^n 
            \prod_{(u,v) \in d} \frac{\gamma_v}{\sum_{v \in \CB} \gamma_v}
        &= \underbrace{e^{-1-z} \left(\sum_{v \in \CB} \gamma_v\right)^n}_{(A)} 
            \cdot \prod_{(u,v) \in d} \rho_v
        &\propto \prod_{v \in \CB} \left( \rho_v \right)^{\text{deg}_d(v)}
    \end{align*}
    We let $\text{deg}_d(v)$ denote the degree of vertex $v \in \CB$ in the assignment $d$.
    The final proportionality claim holds because (A) is fixed and takes the same value for any $d$.
    We can interpret this as saying that the probability that $P^*$ selects a certain assignment is exactly proportional to the product of the normalized dual variables $\rho_v$ for the fully labeled datasets $v \in \CB$ present in that assignment.

    Using the optimal dual variables of the maximum entropy convex program, we have effectively defined a prior distribution $\rho$ on hypotheses as $\rho_v = \gamma_v / \sum_{v \in \CB} \gamma_v$ for all $v \in \CB$.

    We will argue that the optimal (randomized) learner implied by $P^*$ takes the following special form.
    Consider the distribution over assignments $P'$ generated by the following random process where 
    for each $u \in \CA$, we sample an incident $v \in \CB$, independently and with probability proportional to $\rho_v$.
    Then, for any assignment $d \in \CD$, the probability of observing $d$ under the random process is given by the following.
    \begin{align*}
        p'_d = \prod_{(u,v) \in d} \frac{\rho_v}{\sum_{(u, v') \in E} \rho_{v'}} 
        = \prod_{v \in \CB} \left( \rho_v\right)^{\text{deg}_d(v)} \underbrace{\prod_{u \in \CA} \frac{1}{\sum_{(u, v') \in E} \rho_{v'}}}_{\text{(B)}}
        \propto \prod_{v \in \CB} \left( \rho_v\right)^{\text{deg}_d(v)}
    \end{align*}
    Where the proportional follows from the fact that (B) is identical for all assignments $d$.
    Notice that $p^*_d$ and $p'_d$ are proportional to the same quantity for all $d \in \CD$, and therefore the distributions $P'$ and $P^*$ are identical.
    This shows that the optimal randomized learner is in reality, for a given partially labeled dataset $u$, sampling from the previously defined prior over hypotheses, subject to a restriction to only the consistent hypotheses. 
    
    Since each hypothesis consistent with $u$ gives rise to $u$ with equal probability $1/n$, a simple application of Bayes' theorem implies that the optimal learner is sampling from the posterior  $\rho' = \rho | u$ induced by the partially labeled dataset $u$.
\end{proof}

\subsection{Proof and Discussion of Corollary~\ref{corr:randomized_entropy_regularizer}}
\label{appx:proof_randomized_entropy_regularizer}
We now argue that the maximum entropy randomized learner can be viewed as an SRM, and also as an instantiation of the principle of maximum entropy. Recall that for two distributions $P$ and $Q$ supported on a finite set $\Omega$, the \textit{relative entropy} from $Q$ \emph{to} $P$ is defined as
\begin{align*}
    D_{KL} (P \mid Q) = \sum_{x \in \Omega} P(x) \log\left( \frac{P(x)}{Q(x)} \right).
\end{align*}

\begin{lemma}\label{lemma:min_rel_ent}
    Given a distribution $Q$ with finite support $\Omega$, and a subset $\Omega' \subset \Omega$, the distribution $P$ supported on $\Omega'$ which has minimum relative entropy from $Q$ is exactly the restriction of $Q$ to $\Omega'$. 
\end{lemma}
\begin{proof}
    Consider the following optimization problem describing the distribution with minimum relative entropy from $Q$, subject to the constraint that it only have non-zero support on elements of $\Omega'$.
    \begin{align*}
        \min_{P \in \Delta_{\Omega'}} \sum_{x \in \Omega'} P(x) \log\left(\frac{P(x)}{Q(x)} \right)
    \end{align*}
    The partial derivative of the objective with respect to $P(x)$ is $\log\left(\frac{P(x)}{Q(x)}\right) + 1$. Given the constraint to probability distributions on $\Omega'$, the KKT conditions state that all partial derivatives are equal. The unique distribution satisfying the KKT conditions is the restriction of $Q$ to $\Omega'$. 
\end{proof}

We are now equipped to prove Corollary~\ref{corr:randomized_entropy_regularizer}. 

\begin{proof}
Using Lemma~\ref{lemma:min_rel_ent}, we have that when the ground truth for $n-1$ datapoints is revealed as some $u^* \in \CA$, the randomized optimal (maximum entropy) learner is Bayesian in that it updates its posterior over hypotheses to have minimum relative entropy to the prior $\rho$, constrained on outputting hypotheses consistent with $u^*$.

Now, consider the learner induced by the following regularizer:
\begin{align*}
    \psi(H, S_{-i}, x) = \frac{1}{K} \tan^{-1}(D_{KL}(H \mid \rho)),
\end{align*}
where $K>0$ is a parameter, and  $\rho$ corresponds to the previously defined re-normalized dual variables of the maximum entropy convex program. As $K$ grows large,  this learner places more relative importance on empirical risk. Recall that our maximum entropy learner minimizes $D_{KL}(H \mid \rho)$ subject to empirical risk equaling exactly zero. (Equivalently, it minimizes $\tan^{-1} D_{KL}(H \mid \rho)$, which has the favorable property of being bounded above.) Therefore, as $K \to \infty$ the output of the learner induced by $\psi$ converges in total variation distance to the output of our maximum entropy learner.
\end{proof}

In addition to being an SRM in the generalized sense just described, our learner can also be interpreted as an instantiation of the maximum entropy principle. In particular, if the prior $\rho$ were uniform, then indeed our learner would sample from the maximum entropy distribution over hypotheses consistent with the supervised training data. More generally, sampling from the distribution which hues most closely to the prior subject to the provided labels --- as measured by relative entropy --- is the natural generalization of the maximum entropy principle to incorporate prior knowledge. In other words, our learner deviates as little as possible from the prior subject to being consistent with the provided labels.

The fact that our learner simultaneously instantiates SRM and the maximum entropy principle may appear counter-intuitive or even contradictory. Indeed, SRM is the embodiment of \textit{Occam's razor}, which prefers the most simple hypotheses consistent with the data. On the other hand, maximizing entropy might appear to be the opposite of this, as high entropy distributions are arguably more ``complex.'' However, taking the perspective of relative entropy from the uniform distribution (or from a high entropy prior $\rho$), the maximum entropy principle can be viewed as making as few assumptions as possible beyond those substantiated by the data. This is, in fact, fully in accordance with the spirit of Occam's razor.

\section{Extension to Agnostic Learning}\label{sec:generalizing_oig}

Our discussion of learning and one-inclusion graphs has thus far been restricted to the realizable case. Indeed, the structure of one-inclusion graphs, and of the transductive learning setting, depends crucially upon the guarantees provided by learning in the realizable case. In the agnostic case, any notions analogous to the OIG or to transductive learning would be obligated to look considerably different: the fully and partially labeled datasets of Figure \ref{figure:bipartite} would no longer be required to agree with a function in $\CH$, and the adversary of Definition \ref{Definition:trans-learning-error} would likewise be permitted to label its datapoints arbitrarily. But how exactly should an \emph{agnostic} one-inclusion graph be defined, and how would its optimal orientations be judged? How would this relate to an agnostic notion of transductive learning, if at all? 

We devote this section precisely to the development such concepts, and introduce an agnostic one-inclusion graph whose optimal orientations --- judged using vertices' outdegrees minus their ``credits" --- correspond precisely to learners attaining optimal agnostic transductive error. We also demonstrate that an agnostic version of the Hall complexity again characterizes the optimal error rates of hypothesis classes exactly, and exhibit one such optimal learner using maximum entropy programs. 

Throughout the section, we employ the 0-1 loss function. 

\subsection{Transductive Learning}

A crucial tool in our endeavor will be the notion of Hamming distance between functions. 

\begin{definition}
Let $S \in (\CX \times \CY)^n$ be a sample and $h \in \CY^\CX$ a hypothesis. The \defn{Hamming distance} between $S$ and $h$, denoted $||S - h||_0$, is the empirical error incurred by $h$ on $S$, i.e., 
\[ ||S - h||_0 = \sum_{i=1}^n \big[ h(x_i) \neq y_i \big]. \]
When $\CH \subseteq \CY^\CX$ is a hypothesis class, the \emph{Hamming distance} between $S$ and $\CH$ is the minimal Hamming distance incurred between $S$ and any $h \in \CH$, i.e., 
\[ ||S - \CH||_0 = \min_{h \in \CH} || S - h ||_0. \]
\end{definition}

We are now equipped to define the problem of transductive learning in the agnostic case. 

\begin{definition}\label{Definition:agnostic-trans-learning-setup}
The \defn{agnostic transductive learning} setting is that in which the following steps take place: 
\begin{enumerate}
  \item An adversary chooses a collection of $n$ labeled datapoints, $S = \big((x_1, y_1), \ldots (x_n, y_n)\big)$.
  \item The unlabeled datapoints in $S$ are revealed to the learner, i.e., the data of $(x_1, \ldots, x_n)$. 
  \item One datapoint $x_i$ is selected uniformly at random from $S$ as the test point. The information of 
  \[ S_{-i} = \big((x_1, y_1), \ldots, (x_{i-1}, y_{i-1}), (x_{i+1}, y_{i+1}), \ldots, (x_n, y_n) \big) \] 
  is revealed to the learner. 
  \item The learner is prompted to predict the label of $x_i$, i.e., $y_i$. 
\end{enumerate}
\end{definition}

\begin{definition}\label{Definition:agnostic-trans-error}
The \defn{agnostic transductive error} incurred by a learner $A$ on the labeled sample $S$ is its expected error over the uniformly random choice of $x_i$, relative to the performance of the best hypothesis in $\CH$, i.e.,
\[ L_{S}^{\Trans} (A) = \left(\frac{1}{n} \sum_{i \in [n]} \left[ A(S_{-i})(x_i) \neq y_i \right] \right) - \frac 1n \| S - \CH \|_0. \]
\end{definition}

Let us briefly justify that Definitions \ref{Definition:agnostic-trans-learning-setup} and \ref{Definition:agnostic-trans-error} are the appropriate generalizations of their realizable counterparts. We impose two key alterations: 
\begin{enumerate}
    \item The data $S$ selected by the adversary is no longer required to be labeled according to an $h \in \CH$. This agrees precisely with the notion of agnostic PAC learning, in which a learner is required to perform well with respect to any distribution $D$ over $\CX \times \CY$. 
    \item The learner is judged based only on its performance relative to the best hypothesis in $\CH$. This again corresponds to the PAC criterion for agnostic learners, in which learners are benchmarked with respect to hypotheses in $\CH$. 
\end{enumerate}

We now define error rates in the standard manner.

\begin{definition}
The \defn{agnostic transductive error rate} incurred by a learner $A$ is the function
\begin{align*}
    \epsilon^{\ag}_{A, \CH}(n) 
    = \max_{S \in (\CX \times \CY)^n} L_{S}^{\Trans} (A).
\end{align*}
The \emph{agnostic transductive error rate} of a hypothesis class $\CH$ is the pointwise minimal error rate enjoyed by any of its learners, i.e., 
\begin{align*}
    \epsilon^{\ag}_{\CH}(n) = \inf_{A} \epsilon^{\ag}_{A, \CH}(n),
\end{align*}
where the infimum ranges over all learners for $\CH$.
\end{definition}

\subsection{Agnostic One-inclusion Graphs}

We now present a simple modification of the one-inclusion graph that captures the problem of transductive learning in the agnostic case. We note that similar techniques were introduced in the work of \cite{long1998complexity} to analyze binary classification for realizable learning with distribution shift. Our analysis, however, applies to multiclass classification in the agnostic case over arbitrary label sets, and demonstrates an equivalence between (optimal) transductive learners and (optimal) orientations of agnostic one-inclusion graphs. We also introduce an \emph{agnostic Hall complexity}, akin to the Hall complexity introduced in Section \ref{sec:Hall}, that exactly characterizes the errors of optimal transductive learners. 

\begin{definition}
\label{Definition:OIG_agnostic}
Let $\CH \subseteq \CY^\CX$ be a hypothesis class.
The \defn{agnostic one-inclusion graph} of $\CH$ with respect to $S \in \CX^n$, denoted $G_{\ag}(\CH|_S)$, is 
the hypergraph given by the following vertex and edge sets:
\begin{itemize}
    \item $V = \CY^n$, one node for each possible labeling of the $n$ datapoints.
    
    \item $E = \bigcup_{i=1}^n \CY^n|_{S_{-i}}$, where $e \in \CY^n|_{S_{-i}}$ is incident to each $v \in \CY^n$ such that $v|_{S_{-i}} = e$.
\end{itemize}
As in Definition~\ref{Definition:OIG_realizable}, we will sometimes write such an edge $e$ as $e_{(g, i)}$, with $g \in \CY^n$ and $i \in [n]$. Under this representation, $e_{(g, i)}$ is incident to all nodes $v \in \CY^n$ such that $g|_{S_{-i}} = v|_{S_{-i}}$. 
\end{definition}

In the setting of binary classification, with $\CY = \{0, 1\}$, $G_{ag}(\CH|_S)$ is simply the $|S|$-dimensional boolean hypercube. Larger label sets $\CY$ give rise to analogues of the boolean hypercube sometimes referred to as \emph{Hamming graphs}  \citep[Section~12.3.1]{brouwer2011spectra}. Though the vertex and edge sets of the agnostic one-inclusion graph $G_{\ag}(\CH|_S)$ do not explicitly depend on the class $\CH$ itself --- only $S$ --- we will think of $G_{\ag}(\CH|_S)$ as containing the information of which vertices $v \in V$ are members of $\CH|_S$ and which are not. This will be useful information to retain when handling such graphs and, for instance, allows one to deduce $||v - \CH||_0$ for any vertex $v$ using only the information in $G_{\ag}(\CH|_S)$. 

\begin{lemma}
\label{Lemma:agnostic_outdeg_is_trans_error}
Let $\CH$ be a hypothesis class, $S \in (\CX \times \CY)^n$ a sample, and $S_{\CX} \in \CX^n$ the sequence of unlabeled datapoints in $S$. Then the following conditions are equivalent for any learner $A$. 
\begin{enumerate}
    \item $A$ incurs agnostic transductive error at most  $\epsilon$ on the instance $S$. 
    \item $S$, thought of as a vertex in $G_{\ag}(\CH|_{S_\CX})$, has outdegree at most $n \cdot \epsilon + \| S - \CH \|_0$ in the graph oriented by $A$. 
\end{enumerate}
\end{lemma}
\begin{proof}
Let $v \in V$ be the node of $S$ in $G_{\ag}(\CH|_S)$. For an edge $e$ incident to $g$, we write $e \to g$ if $e$ is oriented towards $g$ (by the action of $A$) and $e \not \to g$ otherwise. Then, 
\begin{align*}
    L_{S}^{\Trans} (A) &= - \frac1n \cdot \| S - \CH \|_0 + \frac{1}{n} \sum_{i \in [n]} \left [  A(S_{-i})(x_i) \neq y_i \right ]\\
    &= - \frac1n \cdot \| S - \CH \|_0 + \frac{1}{n} \sum_{i \in [n]} [e_{(g, i)} \not \to g] \\
    &= - \frac1n \cdot \| S - \CH \|_0 + \frac{1}{n} \cdot \outDeg(g).
\end{align*}
\end{proof}

The correspondence between (agnostic) transductive error and vertices' outdegrees again justifies a focus on orientations of one-inclusion graphs that control nodes' outdegrees. Note, however, that degree requirements should no longer be uniform, as they were in Definition \ref{Definition:alpha-(co)orientable}. Each node is instead judged on the basis of its outdegree minus the \emph{credits} it receives as compensation for being distant from $\CH$. This is formalized by the following definition.

\begin{definition}
Let $\CH \subseteq \CY^\CX$ be a hypothesis class and $S \in \CX^n$. We say that $G_{\ag}(\CH|_S)$ is \defn{$(\alpha, \CH)$-agnostic-orientable} if it can be oriented so that all its vertices $v$ have indegrees at least $\alpha - \| v - \CH \|_0$.
Similarly, $G$ is \defn{$(\alpha, \CH)$-agnostic-coorientable} if it can be oriented so that all its vertices $v$ have outdegrees at most $\alpha + \|v - \CH \|_0$. 
We will suppress $\CH$ when it is clear from context, and write simply $\alpha$-agnostic-(co)orientable.
\end{definition}

\begin{lemma}\label{Lemma:agnostic-trans_error_coorientations}
Let $A$ be a learner for $\CH$. The following conditions are equivalent.
\begin{enumerate}
    \item $A$ incurs agnostic transductive error at most $\epsilon$ on all samples of size $n$.
    \item For each $h \in \CH$ and $S \in \CX^n$, $A$ induces an $(n\epsilon)$-agnostic-coorientation on $G_{\ag}(\CH|_S)$.   
    \item For each $h \in \CH$ and $S \in \CX^n$, $A$ induces an $((1 - \epsilon)\cdot n)$-agnostic-orientation on $G_{\ag}(\CH|_S)$. 
\end{enumerate}
\end{lemma}
\begin{proof}
Conditions (2.) and (3.) are  equivalent as a consequence of each node in the undirected graph $G_{\ag}(\CH|_S)$ having degree exactly $n = |S|$, one for each point in $S$ that can be omitted. 
In particular, for any fixed orientation of $G_{\ag}(\CH|_S)$ we have: 
\begin{align*}
    (2.) & \hspace{0.3 cm} \equiv \hspace{0.3 cm} \text{outDeg}(v) \leq n \cdot \epsilon +  \| v - \CH \|_0  \\
    &\iff n - \text{inDeg}(v) \leq n \cdot \epsilon +  \| v - \CH \|_0\\ 
    &\iff \text{inDeg}(v) \geq n(1 - \epsilon) - \| v - \CH \|_0 \\
    &\hspace{0.3 cm} \equiv \hspace{0.3 cm} (3.).  
\end{align*}
The equivalence between (1.)\ and (2.)\ follows immediately from Lemma~\ref{Lemma:agnostic_outdeg_is_trans_error}, applied pointwise to all $S \in (\CX \times \CY)^n$. 
\end{proof}

We now generalize the Hall density and Hall complexity of Section \ref{sec:Hall}, describing agnostic analogues that provide an exact combinatorial characterization of the optimal agnostic transductive error that can be attained on samples of size $n$. (That is, of the transductive error rate of $\CH$.) The central ingredient is to incorporate the non-uniform degree requirements into the Hall arguments of Section \ref{sec:Hall}. Crucially, Hall's theorem and its generalizations are robust to non-uniform degree requirements, allowing us to transfer our reasoning from Section \ref{sec:Hall} in a relatively routine manner. 

\begin{definition}
Let $\CH \subseteq \CY^\CX$ be a hypothesis class, $S \in \CX^n$ a sequence of unlabeled datapoints, and $G_{\ag}(\CH|_S) = (V, E)$ the agnostic one-inclusion graph of $\CH$ with respect to $S$. For a set of nodes $U \subseteq V$, let $E[U] \subseteq E$ denote the collection of edges with at least one incident node in $U$. The \defn{agnostic Hall density} of $G_{\ag}(\CH|_S)$ is
\begin{align*}
\Hall^{\ag}(G) &= \inf_{\substack{U \subseteq V, \\ |U| < \infty }} \frac{|E[U]| + \| U - \CH \|_0 }{|U|}  \\ 
&:=\inf_{\substack{U \subseteq V, \\ |U| < \infty }} \frac{|E[U]| +  \sum_{u \in U} ||u - \CH||_0}{|U|} . 
\end{align*}
\end{definition}

\begin{proposition}\label{Proposition:agnostic_Hall_dimn_orientations}
Let $\CH \subseteq \CY^\CX$ be a hypothesis class, $S \in \CX^n$ a sequence of unlabeled datapoints, and $G_{\ag}(\CH|_S)$ the agnostic one-inclusion graph of $\CH$ on $S$. Then $\Hall^{\ag}\big(G_{\ag}(\CH|_S) \big)$ is the greatest $\alpha$ for which $G_{\ag}(\CH|_S)$ is $\alpha$-agnostic-orientable. 
\end{proposition}
\begin{proof}
The claim follows immediately from the proof of Proposition~\ref{Proposition:Hall_dimn_orientations} and the observation that Hall's theorem is robust to non-uniform degree requirements. That is, an $\alpha$-agnostic-orientation of $G_{\ag}(\CH|_S) = (V, E)$ is a function  $E \to V$ such that each edge is assigned to an incident node and each $v \in V$ receives at least $\alpha - ||v - \CH||_0$ edges. In order for such a function to exist, it is clearly necessary that for each finite $U$, $|E[U]| \geq |U| \cdot \alpha - ||U - \CH||_0$. The classical statement of Hall's theorem --- along with a cloning argument to handle non-uniform degree requirements ---  demonstrates sufficiency when $G_{\ag}(\CH|_S)$ is finite \citep{Hall1935original}. When $G_{\ag}(\CH|_S)$ is infinite, sufficiency follows from the generalization of Hall's theorem to infinite bipartite graphs in which all nodes on the right side have finite degree (and the recollection that each node in $G_{\ag}(\CH|_S)$ has degree $|S|$) \citep{hall1948distinct}. See the proof of 
Proposition~\ref{Proposition:Hall_dimn_orientations} for further detail on the splitting argument. 
\end{proof}

\begin{definition}
\label{appx:agnostic_hall_complexity}
    The \edefn{agnostic Hall complexity} of a hypothesis class $\CH$ is the function $\pi^{\ag}_{\CH} : \N \to \N$ defined: 
    \begin{align*}
        \pi^{\ag}_\CH(n) = \max_{S \in (\CX \times \CY)^n}  n -  \Hall^{\ag}(G_{\ag}(\CH|_S)).
    \end{align*}
\end{definition}

\noindent We now demonstrate that the agnostic Hall complexity exactly characterizes the agnostic transductive error rate, i.e., the transductive error attained by an optimal learner.

\begin{proposition} \label{Proposition:agnostic_Hall_seq_main_result}
    Fix any domain $\CX$, label set $\CY$, and hypothesis class $\CH \subseteq \CY^\CX$.
    Then $\epsilon^{\ag}_\CH(n) = \frac{\pi^{\ag}_\CH(n)}{n}$ for all $n \in \N$.
\end{proposition}
\begin{proof}
First note that $\pi^{\ag}_\CH(n)$ is the minimal $\alpha$ for which all $\left \{ G(\CH|_S) \right\}_{S \in \CX^n}$ are $\alpha$-coorientable, by Proposition \ref{Proposition:agnostic_Hall_dimn_orientations} and the equivalence between conditions (2.)\ and (3.)\ of Lemma \ref{Lemma:agnostic-trans_error_coorientations}. Invoke the equivalence between conditions (1.)\ and (2.)\ of Lemma \ref{Lemma:agnostic-trans_error_coorientations} to complete the proof. 
\end{proof}

\subsection{Equivalence of Errors and Orienting the Agnostic OIG}
\label{subsec:orienting_agnostic_OIG}

A central feature of our work in the realizable case was the equivalence between high-probability, expected, and transductive errors, as established in Section \ref{sec:error}. In particular, it permitted us to freely restrict focus to one-inclusion graphs and their optimal orientations, which are suited to the minimization of transductive error. For learning in the agnostic case, however, the equivalence between errors is not nearly as tight. Informally, the sensitivity of agnostic errors to multiplicative factors renders ineffective many of the arguments from the realizable case. (E.g., doubling the error of a learner for the realizable case is a benign operation, but lethal for an agnostic learner subject to an additive constraint.) 

Using different arguments, however, we are able to demonstrate that the sample complexities corresponding to the expected and transductive errors differ by a factor of at most $3 / \epsilon$. Let us first state the necessary definitions. 

\begin{definition}
The \edefn{agnostic transductive sample complexity} $m^{\ag}_{\Trans, A} : (0, 1) \to \N$ of a learner $A$ is the function mapping $\delta$ to the minimal $m$ for which $\epsilon^{\ag}_{A, \CH}(m') < \delta$ for all $m' \geq m$. That is, 
\[ m^{\ag}_{\Trans, A}(\delta) = \min \{ m \in \N \st \epsilon^{\ag}_{A, \CH}(m') < \delta, \; \forall m' \geq m\}. \] 
The \emph{agnostic transductive sample complexity} of a hypothesis class $\CH$ is the pointwise minimal sample complexity attained by any of its learners, i.e., 
\[ m^{\ag}_{\Trans, \CH} (\epsilon) = \min_A m^{\ag}_{\Trans, A} (\epsilon), \]
where $A$ ranges over all learners for $\CH$. 
\end{definition}

\begin{proposition}
\label{appx:agnostic_errors_equivalence}
Let $\CX$ be an arbitrary domain, $\CY$ a finite label set, and $\CH \subseteq \CY^\CX$  a hypothesis class. Then $m^{\ag}_{\Exp, \CH}(\epsilon) \leq m^{\ag}_{\Trans, \CH}(\epsilon) \leq  m^{\ag}_{\Exp, \CH}(\epsilon / 2) \cdot \frac 3 \epsilon$. 
\end{proposition}
\begin{proof}
In pursuit of the first inequality, let $n = m^{\ag}_{\Trans, \CH}(\epsilon)$ and let $A$ be a learner for $\CH$ attaining this transductive error guarantee. Fix a distribution $D$ over $\CX \times \CY$. We show that $A$ attains favorable expected error on samples of size $n - 1$. 
\begin{align*}
\E_{S \sim D^{n - 1}} L_D \big( A(S) \big) &= \E_{\substack{S \sim D^{n-1}, \\ (x, y) \sim D}} \big[ A(S)(x) \neq y \big] \\
&= \E_{S \sim D^n} \big[ A(S_{-n})(x_{n}) \neq y_n \big] \\
&= \E_{S \sim D^n} \E_{i \in [n]} \big[ A(S_{-i})(x_i) \neq y_i \big] \\
&= \E_{S \sim D^{n}} \frac 1n \outDeg(S) \\
&\leq \E_{S \sim D^{m}} \bigg( \epsilon + \frac 1n  \min_{h \in \CH} \Big|\Big|S - h|_{S} \Big|\Big|_0 \bigg)\\
&\leq \epsilon + \inf_{h \in \CH} \E_{S \sim D^{m}} \frac 1n \Big|\Big|S - h|_{S} \Big|\Big|_0 \\
&= \epsilon + \inf_{h \in \CH} L_D(h)
\end{align*}
Conversely, let $A$ be a learner attaining agnostic expected error at most $\epsilon$ on samples of size $\geq n$. We will extract from $A$ a learner attaining agnostic transductive error at most $2 \epsilon$ on samples of size $n' = \frac{3n}{\epsilon}$. Fix an $S \in \CX^{n'}$; we will design an $(n' \cdot \epsilon)$-agnostic-orientation of $G_{\ag}(\CH|_S) = (V, E)$. In fact, we will design a fractional orientation, i.e., an assignment from each edge $e \in E$ to several of its incident vertices, in non-negative amounts summing to 1. 

Let $e$ be incident to vertices $V_e = \{v_1, \ldots, v_k\}$. Let $x' \in \CX$ be the unique datapoint on which the vertices $V_e$ disagree, when thought of as functions $S \to \CY$. Furthermore, let $D_{v_i}$ be the uniform distribution over the entries of $v_i$, thought of as a sequence of labeled datapoints. For vertices $v_i, v_j$ and $S \sim D_{v_i}$, let $E_{x'}$ denote the event that $S$ does not contain $x'$. 

We now define our fractional orientation by assigning $p_i$ units of $e$ to $v_i$, where 
\[ p_i = \P_{S \sim D_{v_i}^{n}} \big( A(S)(x') = v_i(x') \; \big| \; E_{x'} \big).  \]
Note that $\sum_{i} p_i = 1$ as a consequence of the fact that $A(S)(x') \in \CY = \{v_1(x'), \ldots, v_k(x') \}$. Now fix an arbitrary vertex $v \in G_{\ag}(\CH|_S)$. Let $N(v)$ denote the set of edges incident to $v$, corresponding to some datapoint $x'_{e}$ on which $v$ disagrees with the other nodes incident to $e$. 
Then, 
\begin{align*}
\frac{\outDeg(v)}{n}  &= \frac 1n \sum_{e \in N(v)} \P_{S \sim D_v^n} \left( A(S)(x'_e) \neq v(x'_e) \; \big| \; E_{x'_e} \right)  \\ 
&\leq \frac{1}{\P(E_{x'_e})} \cdot \frac 1n \sum_{e \in N(v)} \P_{S \sim D_v^n} \big( A(S)(x'_e) \neq v(x'_e) \big)  \\ 
&\leq \left( 1 + \frac{\epsilon}{2} \right) \cdot \E_{S \sim D_v^{n}} L_{D_v} \big( A(S) \big) \\
&\leq \left( 1 + \frac{\epsilon}{2} \right) \cdot \left( \epsilon + \inf_{h \in \CH} L_{D_v} (h) \right) \\
&\leq 2 \epsilon + \inf_{h \in \CH} L_{D_v} (h)  \\
&= 2 \epsilon + \inf_{h \in \CH} || h - v ||_0. 
\end{align*}
In the third line, we make use of the fact that 
\[ \P_{S \sim D_v^n} (E_{x'_e}) = \left(1 - \frac{1}{n'} \right)^n \geq 1 - \frac{n}{n'} \geq 1 - \frac \epsilon 3.   \]
The second-to-last line uses the fact that our loss function is bounded above by 1, and thus $\inf_{h \in \CH} L_{D_v} (h)$ as well.
\end{proof}

Given the importance of agnostic OIGs, it is once again natural to ask for characterizations of --- and computational insights on ---  their optimals orientations. We now demonstrate that the maximum entropy learner introduced in Section~\ref{Subsection:Random_SRM_max_ent} generalizes to the agnostic case with analagous guarantees. 

\begin{proposition}
\label{prop:agnostic_dual_characterization}
Let $\CX$ be a domain, $\CY$ a finite label set, and $\CH \subseteq \CY^\CX$ a hypothesis class. Recall the optimal agnostic transductive error rate $\epsilon^{\ag}_{\CH}(n) = \frac{\pi^{\ag}_\CH(n)}{n}$. For any $S \in \CX^n$, let $G_{\ag}(\CH|_S)$ be the agnostic one-inclusion graph of $\CH$ on $S$ and $\CD$ the collection of all orientations in $G_{\ag}(\CH|_S)$ (equivalently, the collection of assignments in the bipartite analogue $G^{\ag}_{\bp} = (\CA, \CB, E)$, where $\CA$ denotes partially labeled datasets and $\CB$ fully labeled datasets\footnote{ I.e., the edge-vertex incidence graph of $G_{\ag}(\CH|_S)$. See the analagous Definition~\ref{defn:bipartite_oig}}.). 

\vspace{0.2 cm}

\noindent Then there is a unique distribution over assignments $P^* \in \Delta_\CD$ such that each $v \in \CB$ receives at least $n - \pi^{\ag}_\CH(n) - \|v - \CH\|_0$ in-degree in expectation, and the following conditions hold simultaneously:
    \begin{enumerate}
        \item $P^*$ achieves maximal entropy among all $P \in \Delta_\CD$ satisfying the in-degree requirements.
        \item $P^*$ corresponds to a randomized transductive learner $\CA$ which, in expectation over its internal randomness, attains optimal error rate $\epsilon^{\ag}_{\CH}(n)$.
        \item $\CA$ can be described as follows:
    \begin{itemize}
        \item Upon receiving the unlabeled datapoints $S_\CX^+ =(x_1,\ldots,x_n)$, including the test point, construct an appropriate prior distribution $\rho$ over $\CH|_{S_\CX^+}$.
        \item Given the index $i$ of the test point, and labels $y_j$ for all datapoints $x_{j \neq i}$, apply a Bayes update to $\rho$ in order obtain a posterior $\rho'$. This posterior corresponds to restricting the prior to hypotheses consistent with the provided labels, and rescaling accordingly.
        \item Sample a hypothesis $h$ from $\rho'$, and output $h(x_i)$ as the prediction for the label of $x_i$.
    \end{itemize}
    \end{enumerate}
\end{proposition}
\begin{proof}
To prove (1.) and (2.), we define a feasible maximum entropy convex program which finds a randomized assignment equivalent to an $(n - \pi^{\ag}_\CH(n))$-agnostic-orientation.
The program is identical to $\maxent$ defined in Proposition~\ref{prop:optimality_of_max_entropy}, but rather than having a fixed $c$ lower bound on the indegree for each fully labeled dataset $v \in \CB$, we have a varying $c_v$ on the RHS constraint in the convex program.
\begin{align*}
    \underset{p_d:\ d \in \CD}{\text{max}} \quad & \sum_{d \in \CD} p_d \ln \frac{1}{p_d} \\
    \text{s.t.} \quad & \sum_{d \in \CD} p_d \sum_{u \in \CA} d(u, v) \geq c_v = n - \pi^{\ag}_\CH(n) - \| v - \CH\|_0
    \quad &\forall v \in \CB \\
    & \sum_{d \in \CD} p_d = 1\\
    & p_d \geq 0 \quad &\forall d \in \CD
\end{align*}
Feasibility of this program holds due to the fact that $\epsilon^{\ag}_\CH(n)$ is the optimal error rate, and that by Lemma~\ref{Lemma:agnostic-trans_error_coorientations} it therefore corresponds to a $(n - \pi^{\ag}_\CH(n))$-agnostic-orientation.
Uniqueness follows from strict concavity of the objective.
These suffice to prove (1.) and (2.).

To address (3.), we need to again take the dual of the program.
Strong duality still holds from the weak version of Slater's condition, as the constraints are still affine.
We omit the full dual derivation as it is straightforward and similar to that of $\maxent$ from the proof of Theorem~\ref{thm:realizable_randomized_learner} in Appendix~\ref{appx:proof_max_entropy}, replacing $c$ with $c_v$ everywhere.
Since $c_v$ falls away when we take the partial derivative w.r.t. $p_d$ or $z$, the rest of the dual derivation is identical.
Therefore, (3.) follows from the proof of Theorem~\ref{thm:realizable_randomized_learner}, which makes no use of $c$.

Note that both the number of edges and nodes will be much larger in the agnostic Hamming graph $G_{\ag}(\CH|_S)$ than the realizable OIG $G(\CH|_S)$.
The number of assignments in their bipartite counterparts will reflect this.
Although the dual of the agnostic and realizable max entropy programs may appear to take the same value for any fixed assignment $d$, since the dual does not have a dependence on $c$, the agnostic max entropy convex program primal actually has a much larger number of primal and dual variables which will impact their associated optimal values.
\end{proof}

Proposition~\ref{prop:agnostic_dual_characterization} demonstrates that the randomized learner from the agnostic case can be viewed as Bayesian, just as the randomized learner from the realizable case.
Furthermore, this indicates the existence of an SRM for randomized agnostic learners, as discussed in the closing discussion of Section~\ref{Subsection:Random_SRM_max_ent} for the realizable setting.
In particular, the associated regularizer tracks KL divergence from the prior $\rho$ defined in (3.)\ of Proposition~\ref{prop:agnostic_dual_characterization}.

\section{Related Work}
\label{appx:related_work}

\paragraph{Learnability and optimal learning rates.} The importance of learners beyond the classical paradigm of empirical risk minimization (ERM) was discussed by \citet{shalev2010learnability}, who exhibit a learnable class that is not learnable by any ERM learner. The authors propose a notion of \emph{stable} learning rules, which they demonstrate characterizes learnability in this setting. Notably, however, they work in Vapnik's general learning setting, which generalizes Valiant's PAC framework in some respects but restricts learners to be proper. Consequently, it would deem as unlearnable such problems as \cite[Theorem 1]{DS14}, a learnable multiclass problem whose learners are all improper. These problems are of central interest to us, and thus the proper learning techniques for achieving stability from \cite{shalev2010learnability} do not resolve our central inquiry.

There is a rich history of related work for one-inclusion graphs in learning, commencing with the seminal work of \citet{haussler1994predicting}, who used the OIG algorithm of \cite{alon1987partitioning} to obtain error guarantees for VC classes.
Subsequently, many follow-up works have studied OIGs and their error.
\citet{rubinstein2009shifting} propose a combinatorial technique called \textit{shifting} to obtain better (sub)graph density bounds for the OIG, and by extension, better mistake bounds.
Within their work, they make use of Hall's theorem, but fall short of characterizing the transductive error exactly with any notion similar to the Hall complexity which we propose in Section~\ref{sec:Hall}.
We note, however, that they were indeed the first to observe the connection between the OIG, the bipartite OIG variant, and Hall's theorem, and as such do not make the claim that our perspective is completely novel.

Our generalized one-inclusion graph proposed in Appendix~\ref{sec:generalizing_oig} is designed to capture learning in the agnostic setting, by extending the vertex set to include all vertices in the \textit{Hamming graph}, not only those labelings of sample points realizable by an $h \in \CH$. This modification closely resembles that proposed by \cite{long1998complexity} for the case of binary classification, though they study realizable learning with distribution shift (rather than learning in the agnostic case). 
\citet{DS14} made several notable advances in the understanding of one-inclusion graphs for multiclass learning, including by exhibiting a learnable problem without any proper learners, improving the analysis of errors incurred by optimal OIG learners, and introducing the DS dimension for measuring the complexity of a hypothesis class. Recently, the breakthrough result of \cite{brukhim2022characterization} used OIGs to prove that the DS dimension indeed characterizes PAC learnability for multiclass classification over arbitrary label sets. They also demonstrate that the related Natarajan dimension, in contrast, does not characterize learnability. 

OIGs are also a key ingredient in the proof of learnability for \emph{partial} concept classes, as studied in \cite{alon2022theory, kalavasis2022multiclass}.
More recently, \citet{OIG-not-always-optimal} demonstrated that optimal orientations of OIGs, despite attaining optimal transductive error, do not necessarily attain optimal PAC error, resolving in the negative a conjecture of \cite{warmuth2004optimal}.
The same authors demonstrate in \cite{aden2023optimal} that although OIGs are not sample optimal in the PAC model, a simple aggregation of OIGs is optimal for multiclass classification over finite label sets. In realizable regression, the recent work of \citet{attias2023optimal} employed OIGs to define the scaled $\gamma$-OIG dimension and demonstrate that it characterizes learnability (unlike the fat shattering dimension). In robust learning, \cite{montasser2022adversarially} designed an optimal learner using their \emph{global one-inclusion graph}. 

\paragraph{Regularization.} Trading off empirical risk with a notion of model complexity harks back to at least the work of \citet{tikhonov1943stability}.
Structural risk minimization, the formalization of this notion within the statistical learning theory community, is usually credited to the celebrated work of \citet{vapnik1974theory}.
There is a large body of work examining how regularizers can impact the speed and stability across learning and optimization (see \cite{zhou2021optimistic, rosset2007L1, lee2010practical, sridharan2008fast} and references therein).
More recently, there is good reason to believe that popular algorithms such as gradient descent, when applied on neural networks, act as implicit (and perhaps \textit{data-dependant}) regularizers \citep{smith2021ImplicitGD, neyshabur2017geometry}.

Our local unsupervised regularizer, however, is unusual in that it can be thought of as an \textit{unsupervised pre-training} algorithm in the transductive setting, which first examines only the unlabeled datapoints in the training set (including the test point), and then uses this to construct a regularizer with which to perform SRM.
The connection between regularization and unsupervised pre-training was proposed at least as far back as \citet{erhan2010does}.
There, the authors demonstrate empirically that in the context of deep learning (as it existed in 2010), unsupervised pre-training can be thought of as an implicit form of regularization through initialization.

Unsupervised pre-training has also seen a reasonable amount of practical success in domains such as computer vision \citep{carreira2016human, chen2017deeplab} and natural language processing \citep{radford2018improving}. 
On the theoretical side, \cite{ge2023provable} perform a study of unsupervised pre-training in which they assume the existence of an underlying latent variable model, and perform maximum likelihood parameter estimation as the unsupervised step. They then perform empirical risk minimization with the pre-trained model.
This setup is somewhat similar in flavor to our algorithms, where the local unsupervised regularizers can be viewed as a form of unsupervised pre-training, and where we perform ERM on the training data plus regularizer.
However, their setup generally differs from ours and they do not focus on characterizing the learnability of multiclass problems.
While there is a modest amount of attention from the community in understanding theoretical properties of unsupervised pre-training as viewed through the lens of self-supervised learning \citep{lee2021predicting, haochen2021provable} --- especially as it relates to language models \citep{saunshi2020mathematical} --- this work usually does not take place in the fundamental setting of supervised multiclass learning. 
Furthermore, unsupervised pre-training usually employs separate datasets for the supervised and unsupervised training phases, whereas our unsupervised regularizer employs the same dataset for both phases of training. 

Perhaps most related to our formalization of regularizers from the perspective of the theory community is the work of \citet{hopkins2022realizable}, who consider the task of extending arbitrary realizable learners into learners for the agnostic case.
In the context of our framework, the extension they provide can be seen as a type of regularization (though not described as so in their work). In particular, their recipe for transforming realizable learners into agnostic learners can be seen as using an unsupervised regularizer in order to restrict focus to a collection of certain hypotheses, on which it then performs empirical risk minimization. Note that restricting focus to certain hypotheses can be implemented as a ``hard" regularizer assigning value $\infty$ to the omitted hypotheses and value zero to the others. This deviates, however, from our setting in several important respects. First, the predictors to which this procedure restricts focus are only elements of $\CH$ if the realizable learner used as input in the reduction is a proper learner. (And, as we have seen, there exist learnable multiclass problems without any proper learners.) Secondly, the technique uses distinct datasets for regularization and minimization of empirical risk, in contrast to our transductive setting. Lastly, and most notably, the result relies on one's being supplied a realizable learner to begin with, whereas we are primarily concerned with the design of learners ``from scratch.''

\end{appendices}

\end{document}